\documentclass[11pt,letterpaper]{article}
\usepackage[letterpaper]{geometry}
\usepackage{algorithm2e}
\usepackage{acl2012}
\usepackage{graphicx}
\usepackage{siunitx}
\usepackage{times}
\usepackage{latexsym}
\usepackage{amsmath}
\usepackage{amssymb}
\usepackage{multirow}
\usepackage{url}
\makeatletter
\newcommand{\@BIBLABEL}{\@emptybiblabel}
\newcommand{\@emptybiblabel}[1]{}
\makeatother
\usepackage[hidelinks]{hyperref}

\usepackage{caption}
\usepackage{subcaption}
\renewcommand{\P}{\ensuremath{\mathbf{P}}}

\newcommand{\E}{\ensuremath{\mathcal{E}}}
\newcommand{\R}{\ensuremath{\mathcal{R}}}
\newcommand{\Normal}{\ensuremath{\mathcal{N}}}

\newcommand{\wtv}{\texttt{word2vec}\xspace}
\newcommand{\sswtv}{\texttt{spacesaving-word2vec}\xspace}

\newcommand{\sslm}{space-saving language model\xspace}
\newcommand{\ssds}{space-saving data structure\xspace}
\newcommand{\ssalg}{space-saving algorithm\xspace}
\newcommand{\SSAlg}{Space-Saving Algorithm\xspace}

\newcommand{\sgns}{SGNS\xspace}

\newcommand{\texteight}{\texttt{text8}\xspace}

\title{Streaming Word Embeddings with the Space-Saving Algorithm}
  
\author{Chandler May, Kevin Duh, Benjamin Van Durme \\
Johns Hopkins University\\
\texttt{cjmay@jhu.edu}\\
\texttt{\{kevinduh,vandurme\}@cs.jhu.edu}
\And
Ashwin Lall\\
Denison University\\
\texttt{lalla@denison.edu}
}

\date{}

\begin{document}

\maketitle

\begin{abstract}
    We develop a streaming (one-pass, bounded-memory) word embedding
    algorithm based on the canonical skip-gram with negative sampling
    algorithm implemented in \wtv.  We compare our streaming algorithm
    to \wtv empirically by measuring the cosine similarity between word
    pairs under each algorithm and by applying each algorithm in the
    downstream task of hashtag prediction on a two-month interval of
    the Twitter sample stream.  We then discuss the results of these
    experiments, concluding they provide partial validation of our
    approach as a streaming replacement for \wtv.  Finally, we discuss
    potential failure modes and suggest directions for future work.
\end{abstract}

\section{Introduction}

Word embedding algorithms such as the skip-gram with negative sampling
(\sgns)~\cite{mikolov2013distributed} method of 
\wtv\footnote{
    \url{https://code.google.com/archive/p/word2vec/}
}
have led to improvements in the performance of many natural language
processing
applications~\cite{turian10word,wang13sequence,socher13parsing,collobert11scratch}.
Existing word embedding training algorithms process the training data in a batch
fashion, passing over it once to estimate a fixed-size vocabulary and
then one or more additional times to learn a word embedding model
based on that vocabulary.  In this study we augment the \wtv word
embedding training algorithm to process training data in a streaming
fashion,
performing all processing in one pass with bounded memory usage.  This
mode of operation facilitates applications in which the data is too
large to store on disk or to scan more than once, and is potentially 
even infinitely large.

We assume the data comprises a stream of sentences
$\left(s_i\right)_i = \left(s_1,s_2, \ldots\right)$. The sentence at
time $t$, denoted $s_t$, consists of a tuple of $n_t$ words
$\left(w^{(t)}_1,w^{(t)}_2,w^{(t)}_3,\ldots,w^{(t)}_{n_t}\right)$;
hereafter we drop the index $t$ for simplicity.  Our
goal is to derive a one-pass, bounded-memory algorithm that, at any
time step $t$, can return word embeddings---vector representations of
words in $\R^D$---that approximately capture the same semantics as word
embeddings learned by a batch algorithm
trained on $\left(s_1,s_2, \ldots, s_{t}\right)$.
Here we defer a rigorous definition of those \emph{semantics} but
illustrate our intent through intrinsic and extrinsic experiments that
measure desirable properties of a streaming word embedding training
algorithm with respect to the corresponding batch algorithm.

Existing word embedding algorithms often use a
first pass through the data to find a $K$-word vocabulary of words
to embed, then use a second pass to learn
the embeddings of those words, ignoring out-of-vocabulary (OOV) words.
Methods for inferring embeddings of OOV words generally
back off to sub-word unit
representations~\cite{luong2013better,dossantos2014,soricut2015,alexandrescu2006,botha2014,lazaridou2013},
but this approach may falter on important classes of words like
named entities.  Conversely, treating words as atomic units
precludes test-time inference of unrecognized words in the
absence of surrounding context.  We do not address this trade-off in
the present study but suggest incorporating sub-word unit
representations into our approach in future work.

In the present study,
we augment \wtv to handle a potentially unbounded vocabulary in bounded
memory using the space-saving algorithm~\cite{metwally2005},
updating an unsmoothed negative sampling distribution online using
reservoir sampling~\cite{vitter1985random}.  We call this augmented,
one-pass, bounded-memory algorithm \sswtv.

\subsection{Incremental SGNS}

The method introduced in this paper, \sswtv, is similar to the recently
proposed \emph{incremental SGNS}~\cite{kaji2017}, developed
independently.  There are three main differences.

First, whereas \sswtv uses the \ssalg to maintain an
approximate vocabulary, incremental SGNS uses the Misra-Gries
algorithm~\cite{misra1982}.
Prior work found the \ssalg to outperform a modern
implementation of the Misra-Gries algorithm~\cite{demaine2002} and
many other streaming frequent-item algorithms in both error and
runtime~\cite{cormode2008}.

Second, whereas \sswtv uses standard
reservoir sampling to estimate an unsmoothed negative sampling
distribution, \newcite{kaji2017} develop a modified reservoir
sampling algorithm to estimate a smoothed negative sampling
distribution.  In prior work, smoothing the negative
sampling distribution was shown to increase word embedding quality
consistently~\cite{levy2015improving}.

Third, whereas \sswtv employs a separate, thresholded, linearly
decaying learning rate for each embedding,
resetting the learning rate for an embedding each time the
corresponding word in the \ssds is replaced,
incremental SGNS employs AdaGrad to adaptively set
per-word learning rates~\cite{duchi2011}.  The
vanishing learning rates estimated by AdaGrad~\cite{zeiler2012} may be
inappropriate in the streaming setting if the data is not i.i.d.\ and
we desire a model that gives similar weight to data from the beginning,
middle, and end of the stream.

In light of the previous discussion, we cast our contribution as a
complementary
implementation and analysis to that of \newcite{kaji2017}.
We also release an open-source C++ implementation of our algorithm in
hopes of enabling comparison and future work.

In what follows, we first review
\wtv, the \ssalg, and reservoir sampling in
Section~\ref{sec:background}.  We then introduce our algorithm,
referred to as \sswtv, in Section~\ref{sec:sswtv}.  In
Section~\ref{sec:intrinsic} we describe and report the results of
intrinsic experiments on word embeddings learned by \wtv and \sswtv;
then, in Section~\ref{sec:extrinsic}, we describe and report on the
results of an application of \wtv and \sswtv in a downstream task of
hashtag prediction on Twitter.  We discuss modeling and
implementation questions along with the results of our experiments in
Section~\ref{sec:discussion}, also offering suggestions for future
work, and conclude in Section~\ref{sec:conclusion}.

\section{Background}
\label{sec:background}

\subsection{\wtv} 

The intuition behind \wtv is to build on the distributional
hypothesis~\cite{harris1954,sahlgren2008} and map words used in
similar contexts to nearby vectors in Euclidean space.  Let $w$ be a
word in the vocabulary and let $v_w \in \R^D$ denote the target-word
embedding of $w$ in Euclidean space.  Let $x$ be a word in a
context of $w$ (for example, a word co-occurring with $w$ in a
sentence) and let $v'_x \in \R^D$ denote the context-word
embedding of $x$.  Starting with a random
initialization of all target-word and context-word embeddings, the
\sgns algorithm of \wtv uses stochastic gradient descent to try to
maximize $\sigma\left(\langle v_w, v'_x \rangle\right)$ for
co-occurring words $w$ and $x$, where $\sigma$ is the sigmoid function,
$\sigma(t) = 1/(1+\exp(-t))$.
Simultaneously, to constrain the problem, \sgns seeks to minimize
$\sigma\left(\langle v_w, v'_y \rangle\right)$ for words $w$ and $y$
that do not co-occur.
Thus, if both $w$ and $z$ occur in similar contexts (for example, both
co-occur with $x$), \wtv will learn similar target-word embeddings for
them.

The complete objective function of \sgns then comprises the sum
over all co-occurring target-word--context-word pairs ($w$, $x$) of the
expression
\begin{align}
    \log \sigma\left(\langle v_w,v'_x\rangle\right) +
    S\, \E_{y \sim \P}\! \left[
        \log \sigma\left(- \langle v_w,v'_y\rangle\right)
    \right]
    \label{eq:sgns}
\end{align}
where $S$ is a fixed pre-specified integer, $\P$ is a
\emph{negative sampling}
distribution over words, and the expectation over $y$ is
estimated by the mean of $S$ i.i.d.\ samples from $\P$.
The negative sampling distribution represents random noise and is
implemented in practice as a smoothed empirical unigram distribution,
where smoothing is accomplished by raising all word probabilities to
the $0.75$ power and re-normalizing.

While Equation~\eqref{eq:sgns}
can be optimized in an online fashion, one word context at a time,
\wtv first scans through the entire data to compute the vocabulary (the
set of all word types occurring in the data, often truncated in
practice to only those words occurring at least five or so times) and
negative sampling distribution.
To augment \wtv to learn word embeddings in a single pass, using
bounded memory, we must develop an approach to handle an unknown
and potentially unbounded vocabulary and estimate a
usable negative sampling distribution $\P$ over it.

\begin{algorithm*}
    \KwData{Stream of sentences $(s_i)_i$,
    vocabulary size $K$, negative sampling reservoir size $N$.}

    \KwResult{Any-time word embeddings indexed against \ssds.}

    initialize empty size-$K$ \ssds and size-$N$ negative sampling
    reservoir \;

    initialize input, output word embeddings \;

    \For{ever}{
        read sentence $s = (w_1, \ldots, w_J)$ (a tuple of words) \;
        subsample sentence $s' = (w'_1, \ldots, w'_{J'})$ \;
        insert words from $s'$ into \ssds and their \ssds
        indices into reservoir \;
        \For{each skip-gram context in $s'$}{
            \If{all words in context are in \ssds}{
                take \sgns gradient steps on \ssds indices
                corresponding to words in this context \;
            }
        }
    }
\caption{The \sswtv algorithm, simplified for ease of exposition.  See
    the text for details, and see Algorithm~\ref{alg:sswtv-full} for
    a full algorithm listing.}
\label{alg:sswtv}
\end{algorithm*}

\subsection{\SSAlg}

The \ssalg is a one-pass algorithm that estimates the most
frequent items in a stream in bounded memory~\cite{metwally2005}.
It does so by maintaining an array of $K$
items (for a pre-specified number $K$) and corresponding array of $K$
counts; when $K$ unique items have
been seen and a new item is encountered, the item with smallest count
is replaced while the corresponding count is incremented.  The \ssalg
thus over-estimates item counts but enjoys a bound of $n/K$ on the
error in each count, where $n$ is the total number of items (including
repeats) seen so far.

Consider a stream of items $(x_i)_i$.  To initialize
the \ssalg, a size-$K$ array of samples $(z_1, \ldots, z_K)$
is initialized with empty values and a
corresponding array of their respective counts $(c_1, \ldots, c_K)$
is initialized with zeroes.  At each time step $i$, item $x_i$ is
observed and one of three actions is taken.
\begin{enumerate}
\item If $x_i$ is in the array $(z_1, \ldots, z_K)$ then increment the
    corresponding count in $(c_1, \ldots, c_K)$ by one.
\item If $x_i$ is not in the array $(z_1, \ldots, z_K)$ but the array
    is not full (we have not seen $K$ unique items yet) then set the
        next empty array slot to $x_i$ and increment the corresponding
        count by one.
\item Otherwise replace the element in $(z_1, \ldots, z_K)$ with
    smallest count with $x_i$ and increment the corresponding count by
        one.
\end{enumerate}
At any time $i$ in the stream the \ssds contains all items with true
count greater than $i/K$ seen so far and over-estimates all counts by
at most $i/K$~\cite{metwally2005}.  The \ssds can be efficiently
implemented with a min-heap, with multiple linked lists, or with
multiple arrays with additional pointers (emulating linked lists for
the special case of the \ssalg).

\subsection{Reservoir Sampling}

Reservoir sampling is a one-pass algorithm that computes a
uniform subsample of a stream in bounded
memory~\cite{vitter1985random}.
To initialize the sampler, an array of $K$ items $(r_1,
\ldots, r_K)$ is initialized with empty values and a
counter $n$ is initialized to zero.  Then, when stream item $x_i$
is seen, $n$ is incremented by one and one of two actions is taken.
\begin{enumerate}
    \item If $i \le K$, set $r_i$ to $x_i$.
    \item Otherwise draw an integer $k$ uniformly from $\{1, 2, \ldots,
        n\}$ and, if $k$ is less than or equal to $K$, replace $r_k$
        with $x_i$.
\end{enumerate}
At any time $i$ in the stream the reservoir contains a uniform sample
of the items seen so far~\cite{vitter1985random}.

\section{\sswtv}
\label{sec:sswtv}

To learn word embeddings in one pass using bounded memory we use the
\ssalg to create and update a vocabulary of $K$ frequent words and we
use reservoir sampling to maintain a negative sampling distribution
over that vocabulary.
Specifically, we start the learning process by initializing all word
embeddings randomly and initializing the \ssds and negative sampling
reservoir as empty.  For each sentence in the data
stream we first subsample
the words in the sentence, following \wtv.  We then insert all retained
word tokens into the \ssds and negative sampling reservoir.  In each
fixed-length context window, we then check if all words
are present in the \ssds.
If so, we iterate over the left and right \emph{context} words of the
central \emph{target} word, performing a gradient step of
Equation~\eqref{eq:sgns} for each context-word--target-word pair in
turn.
Simplified pseudo-code of the \sswtv algorithm is given
in Algorithm~\ref{alg:sswtv}; more detailed pseudo-code is provided
in Algorithm~\ref{alg:sswtv-full}.

While the high-level description of \sswtv is straightforward
there are several important implementation choices; we summarize them
here.  We take care to distinguish between a word $w$ and its index
in the \ssds, which we denote by $k$.  By nature of the \ssalg, a given
index $k$ can be associated with different words at different points
during training.
\begin{itemize}
\item Whenever a word $w$ is ejected from the \ssds its target-word
    embedding $v_k$ and context-word embedding $v'_k$ are
        re-initialized as draws from $\Normal(0, 1)^D$.  This means
        that if $w$ appears in the future and is inserted again into
        \ssds, training of its embedding starts over from scratch.
\item Sentence subsampling retains all words not in the \ssds,
    and retains each word $w$ in the \ssds with probability
        $\min(1, \sqrt{\delta / f_k})$, where $f_k$ is the count of
        \ssds element $k$.  All other words in the sentence are
        discarded.
\item When words are discarded by sentence subsampling, we do not add
        them to the \ssds and negative sampling reservoir.  When a
        sentence smaller (after subsampling) than a single context
        window is encountered, its words are first added to the \ssds
        and negative sampling reservoir, and then the algorithm moves
        on to the next sentence, skipping the embedding gradient
        updates.
\item When a skip-gram context contains an out-of-vocabulary word, the
    entire skip-gram is ignored. (However, for small context window
        sizes we expect this event to occur only rarely, as the \ssds
        empirically has a long tail, and when the out-of-vocabulary
        check is performed all words in the context would have just
        received an incremented count.)
\item Our negative sampling approach
    differs from that of \wtv in that we draw from the empirical
        distribution over words in the stream instead of over a
        smoothed empirical distribution.
\item We maintain a separate, thresholded, linearly decaying
        learning rate for each of the $K$ bins in the \ssds, resetting
        a learning rate when the corresponding word is ejected from the
        \ssds.
\end{itemize}

We leave a thorough investigation of the impacts of these design
choices to future work.
Our C++ code implementing the \sswtv algorithm with these choices
is released as open-source software to facilitate reproducibility and
future work.\footnote{
    \url{https://github.com/cjmay/athena}
}

\section{Intrinsic Evaluation}
\label{sec:intrinsic}

\begin{figure*}
    \scriptsize
    \centering
    \begin{subfigure}[b]{0.59\columnwidth}
    \includegraphics[scale=0.53]{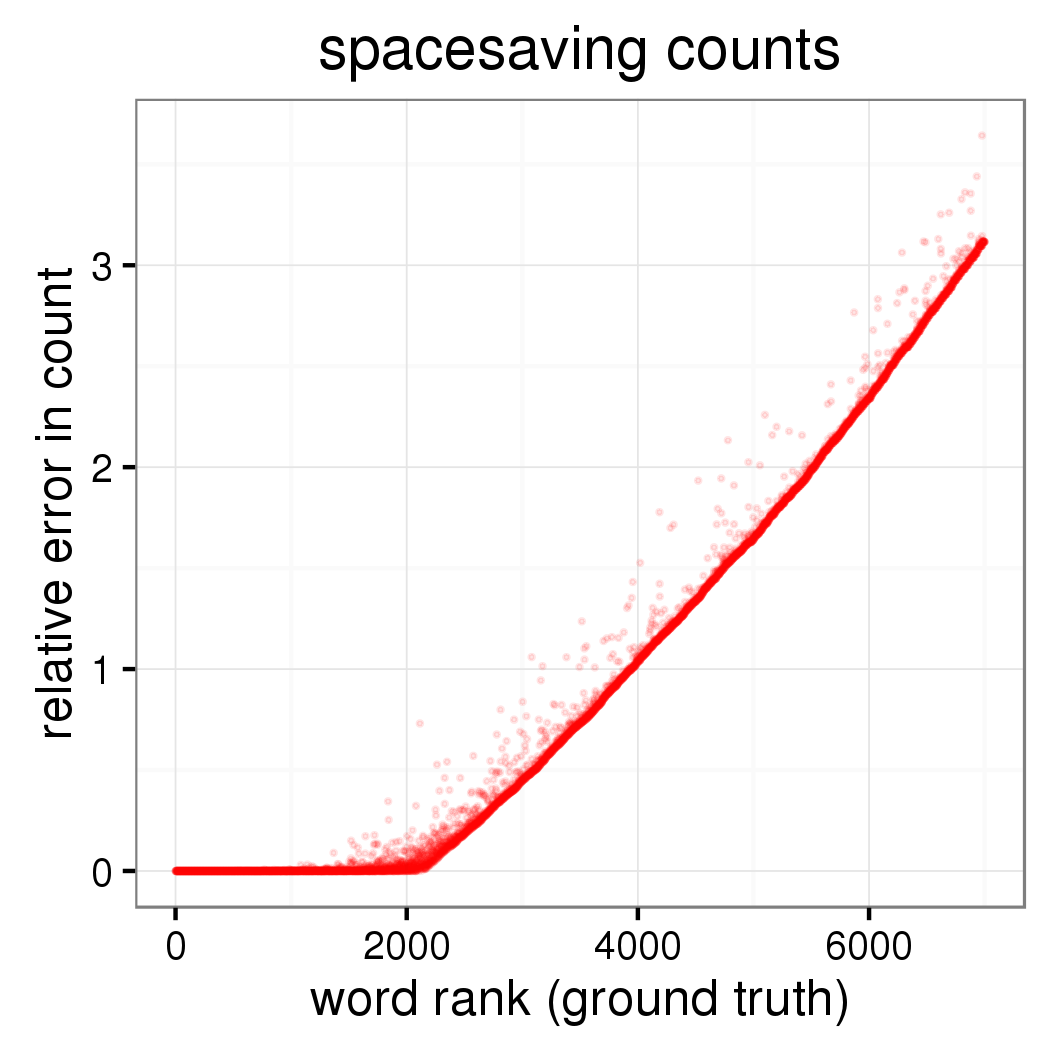}
    \caption{$K=7000$}
    \label{fig:text8-counts-7000}
    \end{subfigure}
    \begin{subfigure}[b]{0.59\columnwidth}
    \includegraphics[scale=0.53]{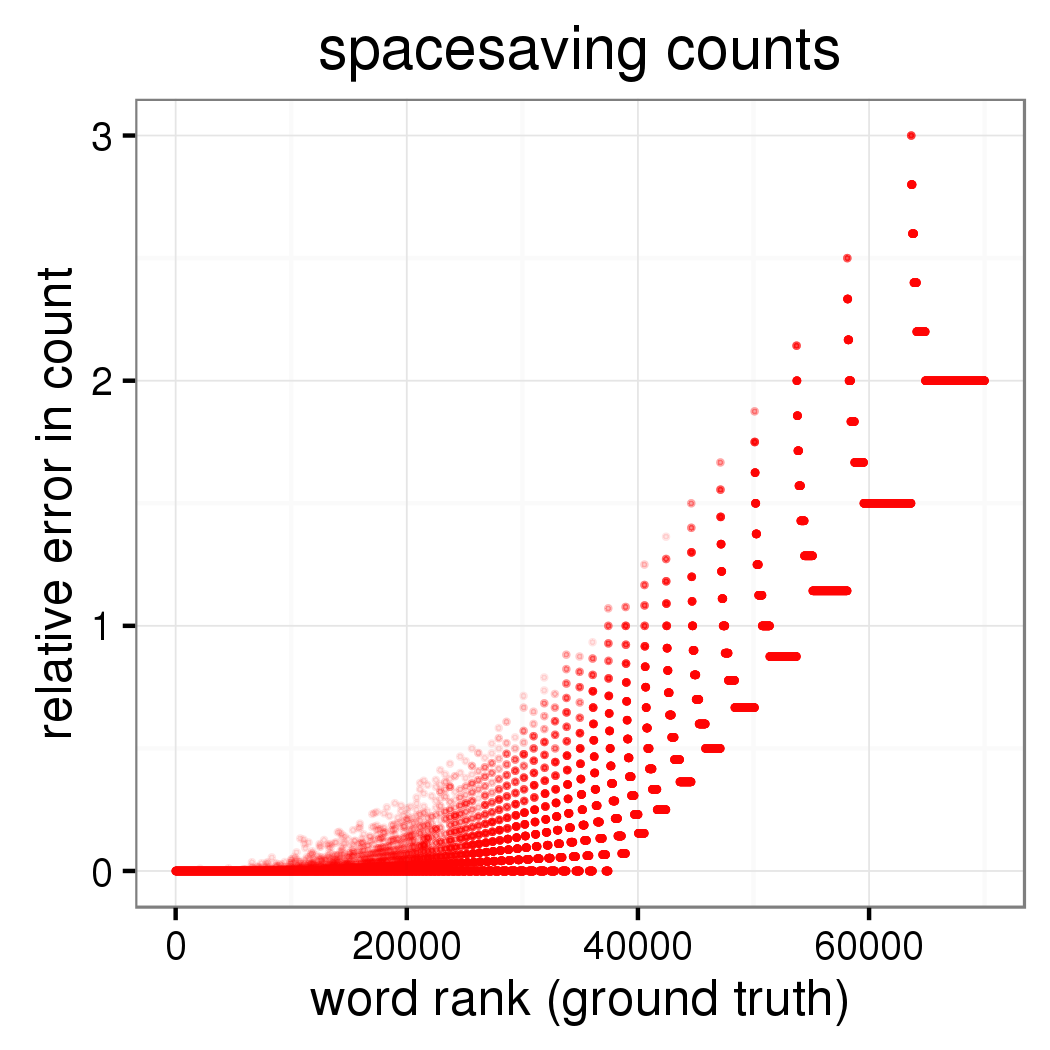}
    \caption{$K=70000$}
    \label{fig:text8-counts-70000}
    \end{subfigure}
    \begin{subfigure}[b]{0.59\columnwidth}
    \includegraphics[scale=0.53]{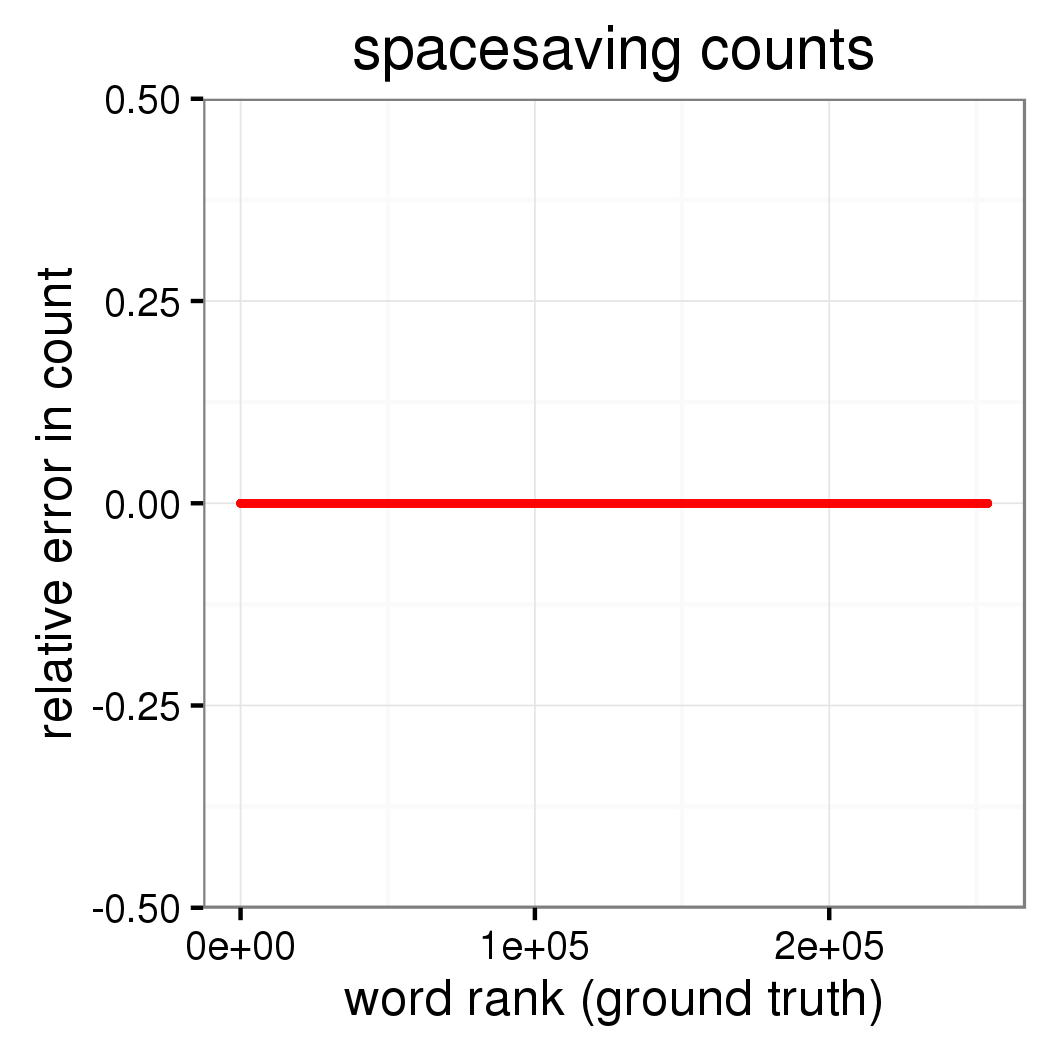}
    \caption{$K=700000$}
    \label{fig:text8-counts-700000}
    \end{subfigure}
    \caption{
        Relative error in space-saving estimates of word counts on the
        \texteight data set, using vocabulary sizes of
        \protect\subref{fig:text8-counts-7000} \num{7000},
        \protect\subref{fig:text8-counts-70000} \num{70000}, and
        \protect\subref{fig:text8-counts-700000} \num{700000},
        imputing counts of words dropped from the \ssds.
    }
    \label{fig:text8-counts}
\end{figure*}

\begin{figure}
    \scriptsize
    \centering
    \includegraphics[scale=0.53]{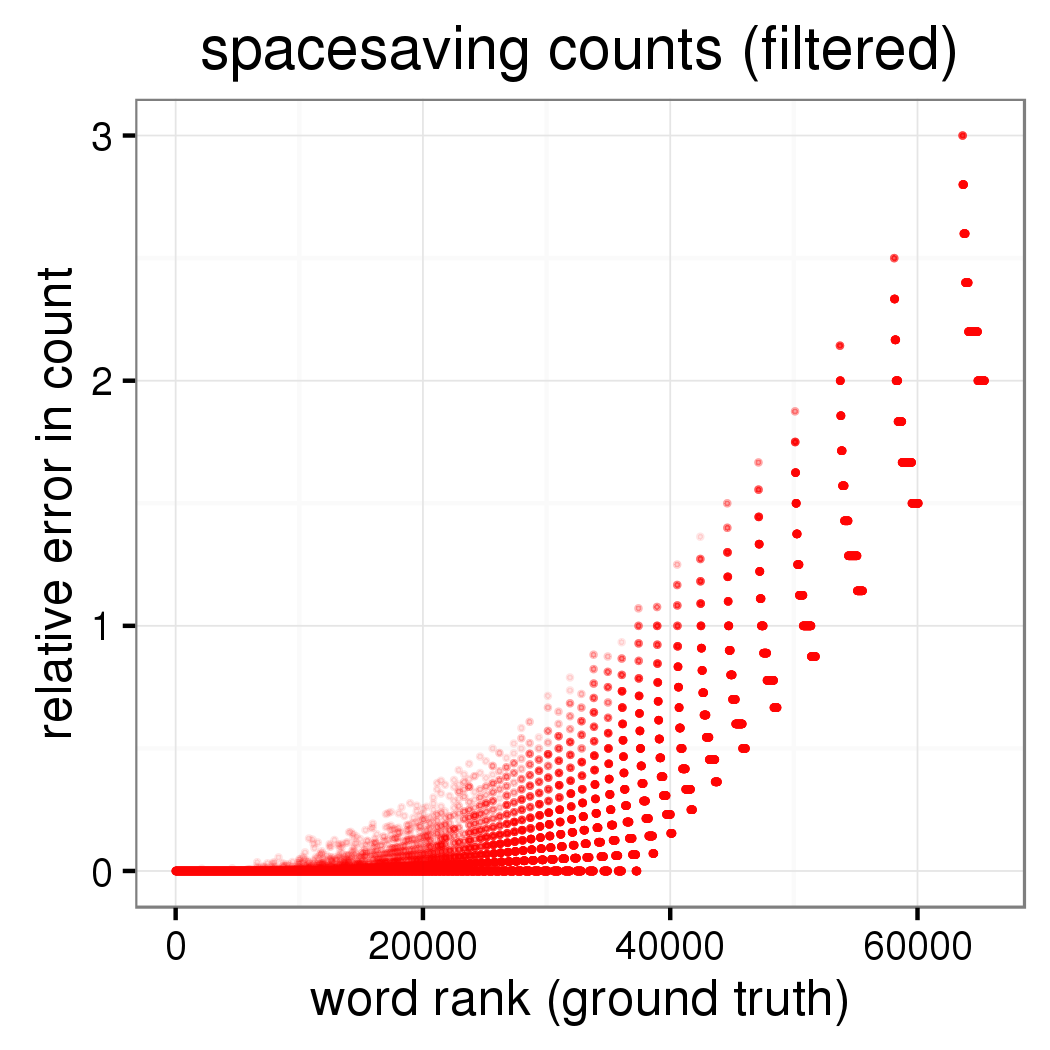}
    \caption{
        Relative error in space-saving estimates of word counts on the
        \texteight data set, using a \num{70000}-dimensional
        vocabulary,
        omitting words dropped from the \ssds rather than imputing
        their counts.
        Vertical strata correspond to words with the same ground-truth
        frequency, hence the same denominator in the relative error of
        their \sslm estimates; the right-most stratum corresponds to
        words with ground-truth frequency five.
    }
    \label{fig:text8-70000-counts-filter-missing}
\end{figure}

To inspect the empirical operation of the \sswtv algorithm,
we perform two intrinsic evaluations.  First, we estimate the
errors in word counts in the \ssds for different \ssds sizes, noting
that those errors affect both the vocabulary and the negative sampling
distribution.  Second, for a large number of word pairs in the shared
vocabulary, we compare the cosine similarity of each pair under \wtv to
the cosine similarity under \sswtv.  The aim of this second experiment
is to measure a rudimentary formulation of the degree to which
\sswtv embeddings approximate \wtv embeddings.

For both intrinsic evaluations, we use \texteight,\footnote{
    \url{http://mattmahoney.net/dc/text8.zip}
}
the first \num{100}~MB of cleaned text from a Wikipedia dump dated
3 March 2006.\footnote{
    \url{http://mattmahoney.net/dc/textdata.html}
}
To estimate the errors in \ssds word counts, we treat the data set as
a stream of words and apply the \ssalg to that stream separately for
\ssds sizes ($K$) of \num{7000}, \num{70000}, and \num{700000},
learning three different sets of approximate word counts.
There are a
total of \num{253854} word types in the \texteight data set, so
vocabularies of size \num{7000} or \num{70000} are truncated and
represent streaming approximations, whereas a vocabulary of size
\num{700000} contains all word types.

For each \ssds size, we compute the true word
counts in the data set and plot the relative error of the approximate
count of each word with respect to its rank in the true ordering by
frequency in the data set (descending).  When computing and plotting
these errors we estimate the count of words dropped from the \ssds
as the value of the smallest counter in the \ssds.
Separate plots for the three
streaming approximations (three values of $K$) are shown in
Figure~\ref{fig:text8-counts}.
For $K=7000$ and $K=70000$, we observe small error for high-frequency
words and a slow increase in error with respect to rank initially,
followed by a distinctive linear increase in error with respect to rank
for lower-frequency words. In both cases the relative error is
around one or two for many words; however, we note there are ten times
as many words portrayed in the $K=70000$ plot, hence the accumulated
error is much smaller.
In the $K=7000$ case the dropped counts lie
on the positive-slope line to the right of the kink; in the
$K=70000$ case the dropped counts are the points displayed as
horizontal lines emanating rightward from a similar positive-slope
line.  The $K=70000$ relative errors, \emph{omitting} dropped words,
are depicted in Figure~\ref{fig:text8-70000-counts-filter-missing} for
the sake of comparison. For $K=700000$ we find zero error for all
word ranks, reflecting the fact that the entire vocabulary now fits in
the \ssds.

\begin{figure}
    \scriptsize
    \centering
    \begin{subfigure}[b]{0.49\columnwidth}
    \includegraphics[scale=0.43]{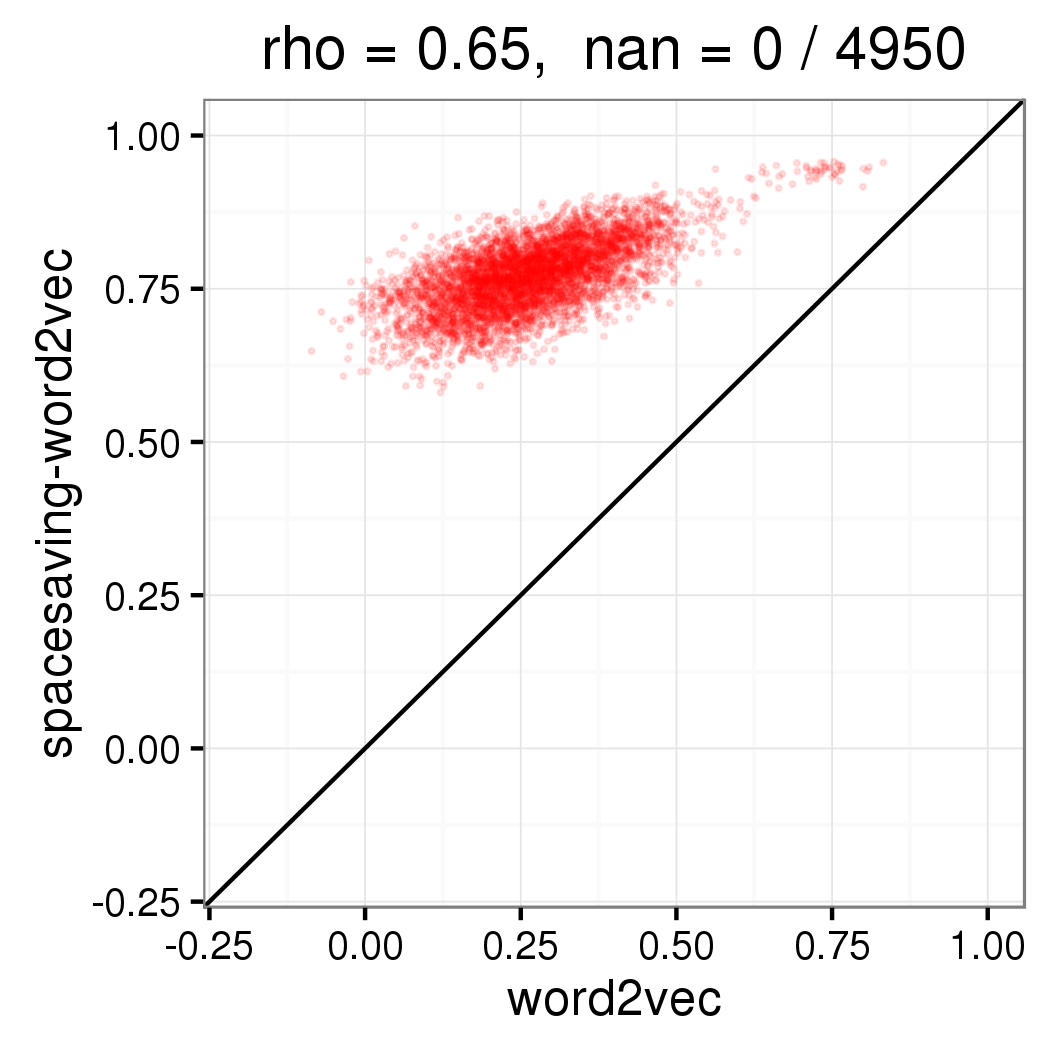}
    \caption{1--100, 1--100}
    \label{fig:text8-7000-1-100-1-100}
    \end{subfigure}
    \begin{subfigure}[b]{0.49\columnwidth}
    \includegraphics[scale=0.43]{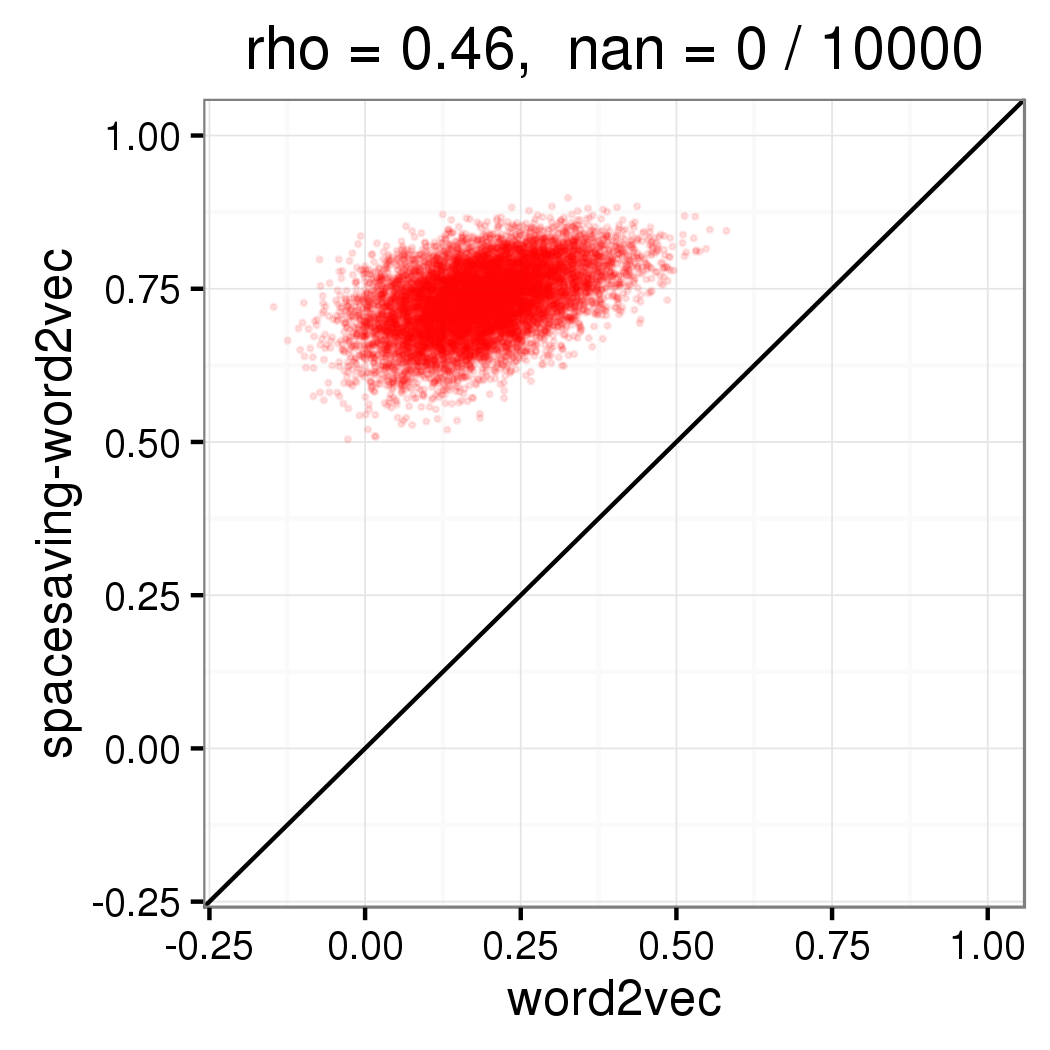}
    \caption{1--100, 1601--1700}
    \label{fig:text8-7000-1-100-1601-1700}
    \end{subfigure}
    \begin{subfigure}[b]{0.49\columnwidth}
    \includegraphics[scale=0.43]{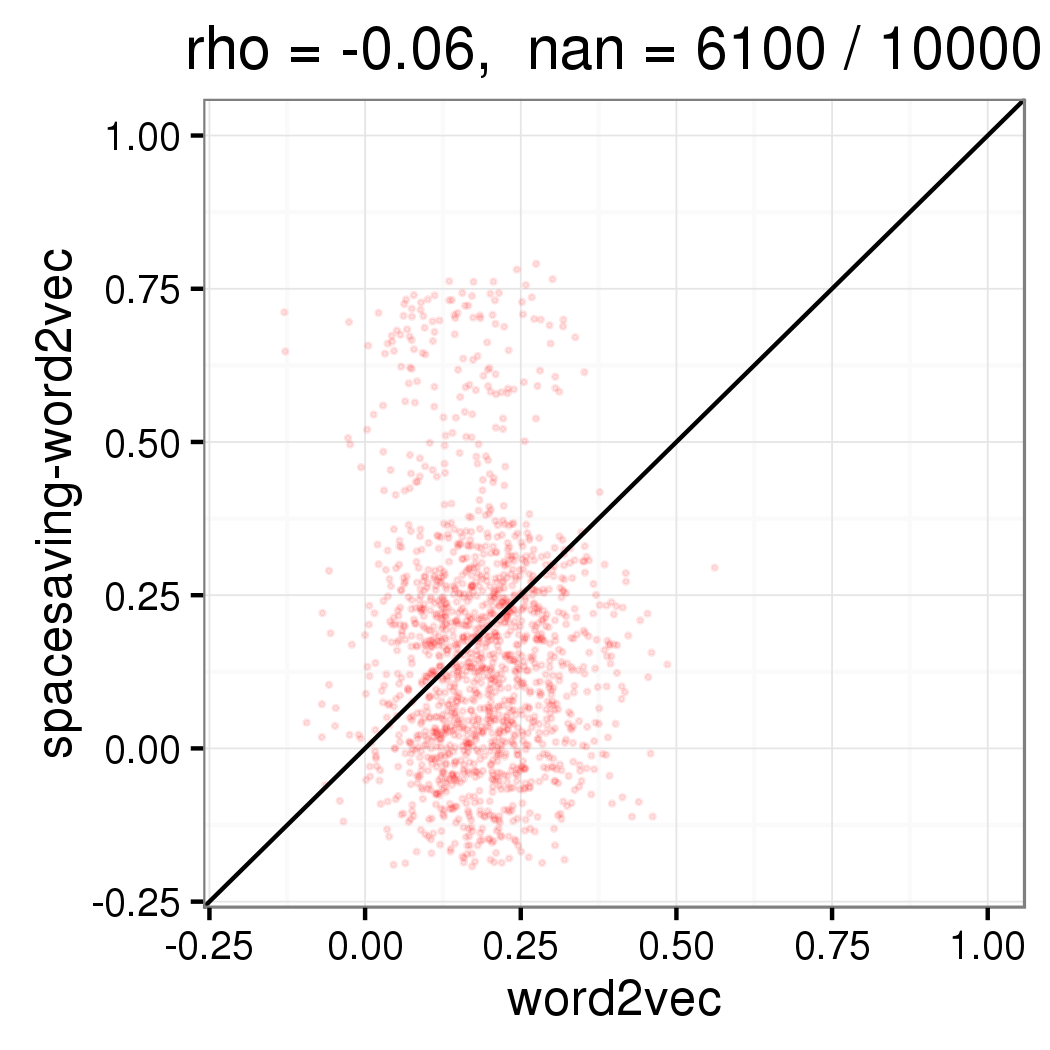}
    \caption{1--100, 6401--6500}
    \end{subfigure}
    \begin{subfigure}[b]{0.49\columnwidth}
    \includegraphics[scale=0.43]{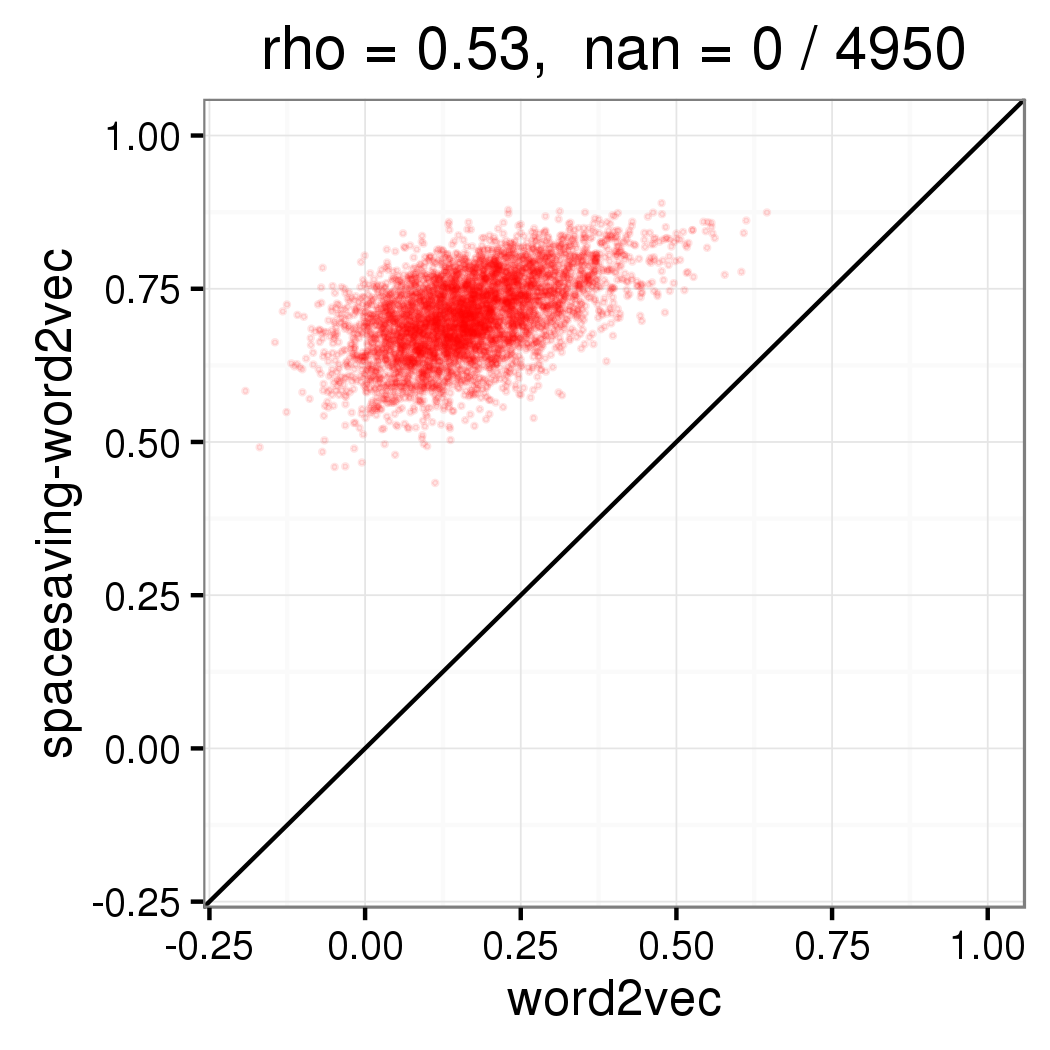}
    \caption{1601--1700, 1601--1700}
    \end{subfigure}
    \begin{subfigure}[b]{0.49\columnwidth}
    \includegraphics[scale=0.43]{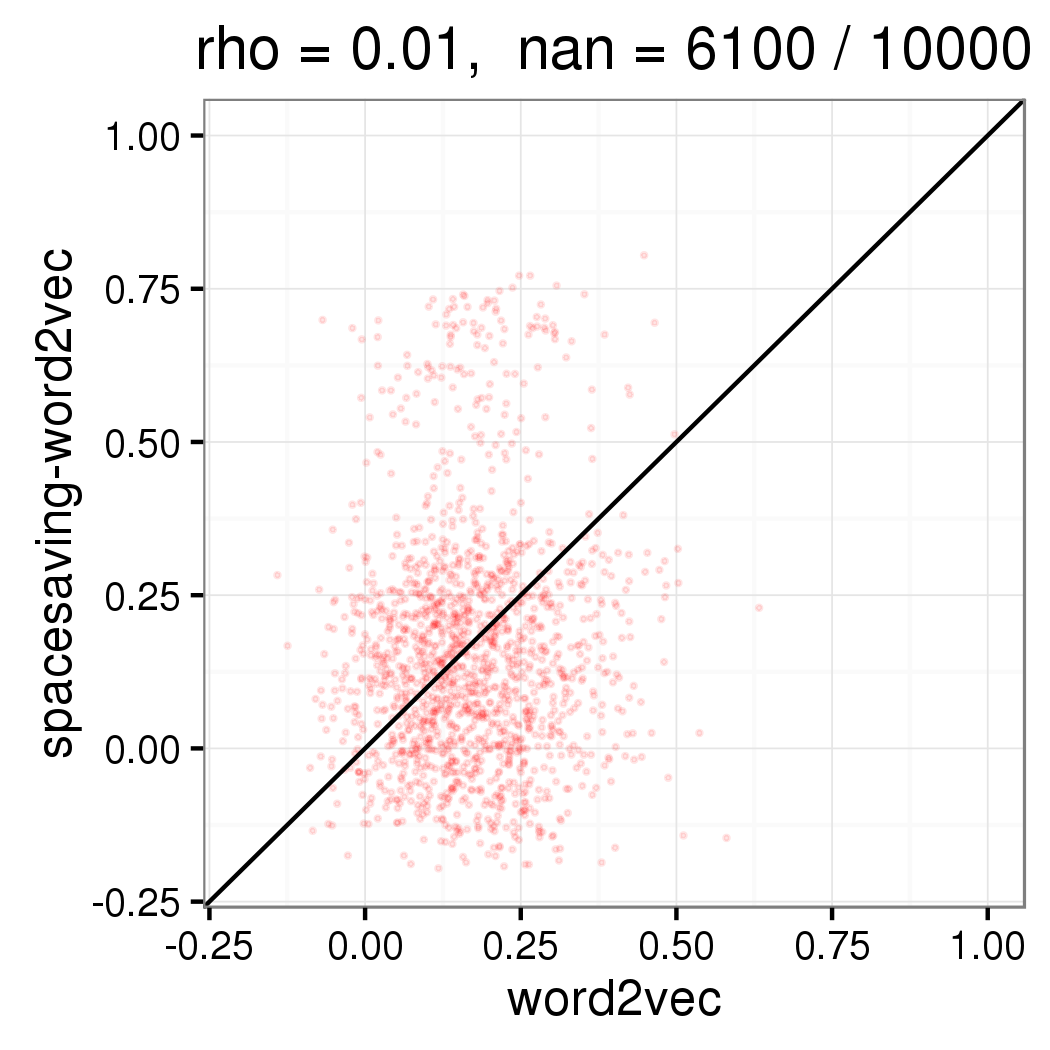}
    \caption{1601--1700, 6401--6500}
    \end{subfigure}
    \begin{subfigure}[b]{0.49\columnwidth}
    \includegraphics[scale=0.43]{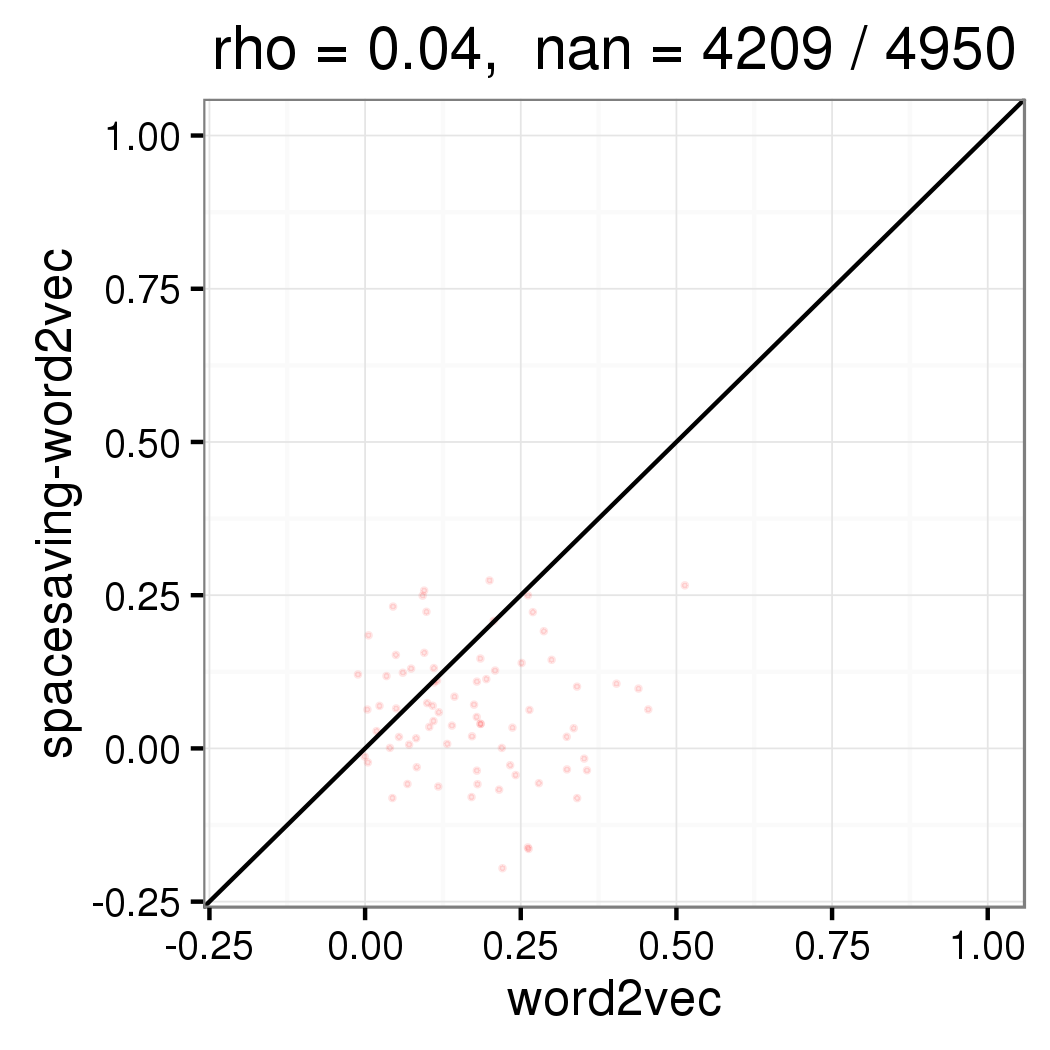}
    \caption{6401--6500, 6401--6500}
    \end{subfigure}
    \caption{Cosine similarity of selected word pairs under \sswtv
    versus \wtv, using \num{7000}-dimensional vocabulary.  Models are
    learned on the \texteight data set.  The word pairs represented in
    each plot are the unique word pairs (up to ordering) in the
    Cartesian product of words in one range of ranks (by frequency)
    with words in another range of the ranks.  For example,
    Figure~\ref{fig:text8-7000-1-100-1-100} depicts unique pairs of
    words in which both words are drawn from the top \num{100} words
    (by frequency), whereas Figure~\ref{fig:text8-7000-1-100-1601-1700}
    depicts pairs of words in which one word is drawn from the top
    \num{100} words and the other is drawn from word ranks \num{1601}
    to \num{1700} (inclusive).}
    \label{fig:text8-7000-sim}
\end{figure}

\begin{figure}
    \scriptsize
    \centering
    \begin{subfigure}[b]{0.49\columnwidth}
    \includegraphics[scale=0.43]{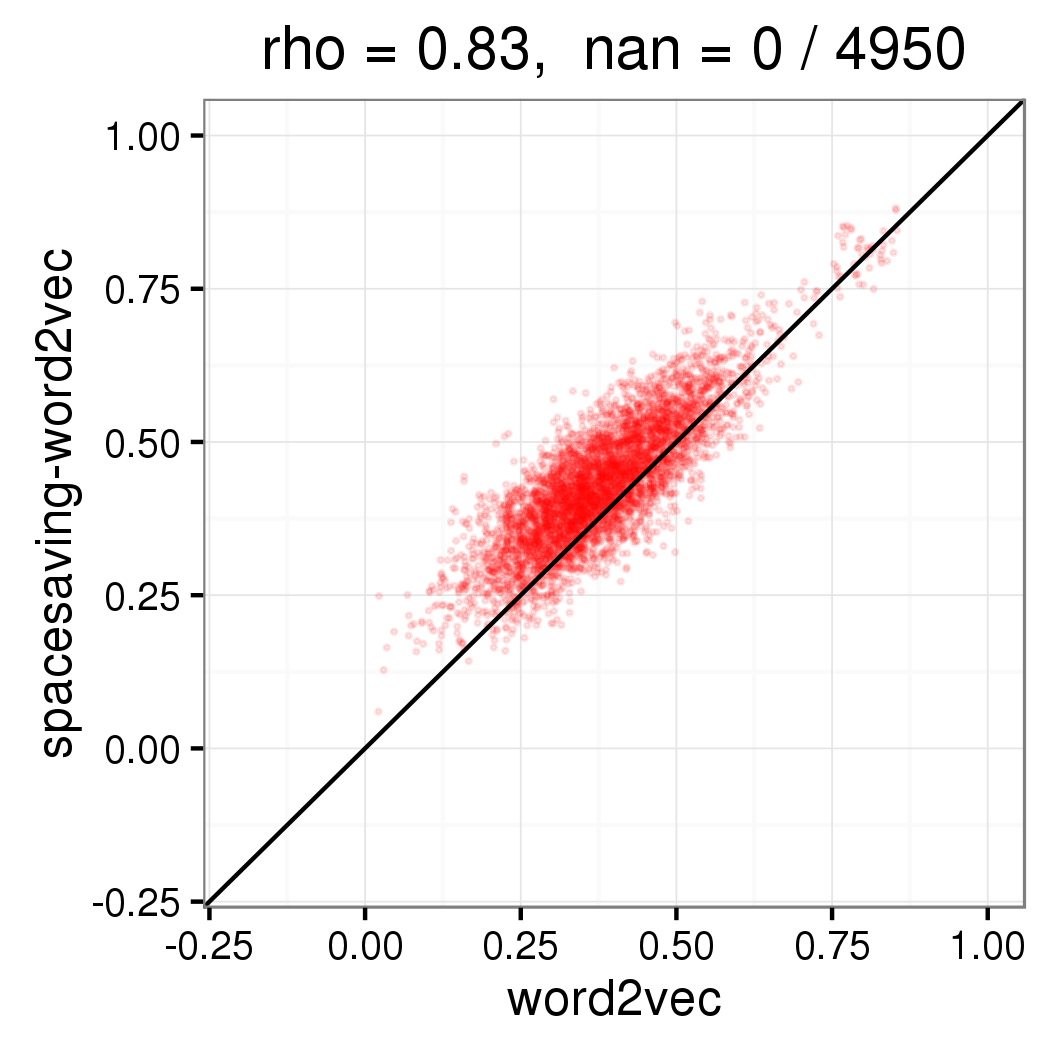}
    \caption{1--100, 1--100}
    \end{subfigure}
    \begin{subfigure}[b]{0.49\columnwidth}
    \includegraphics[scale=0.43]{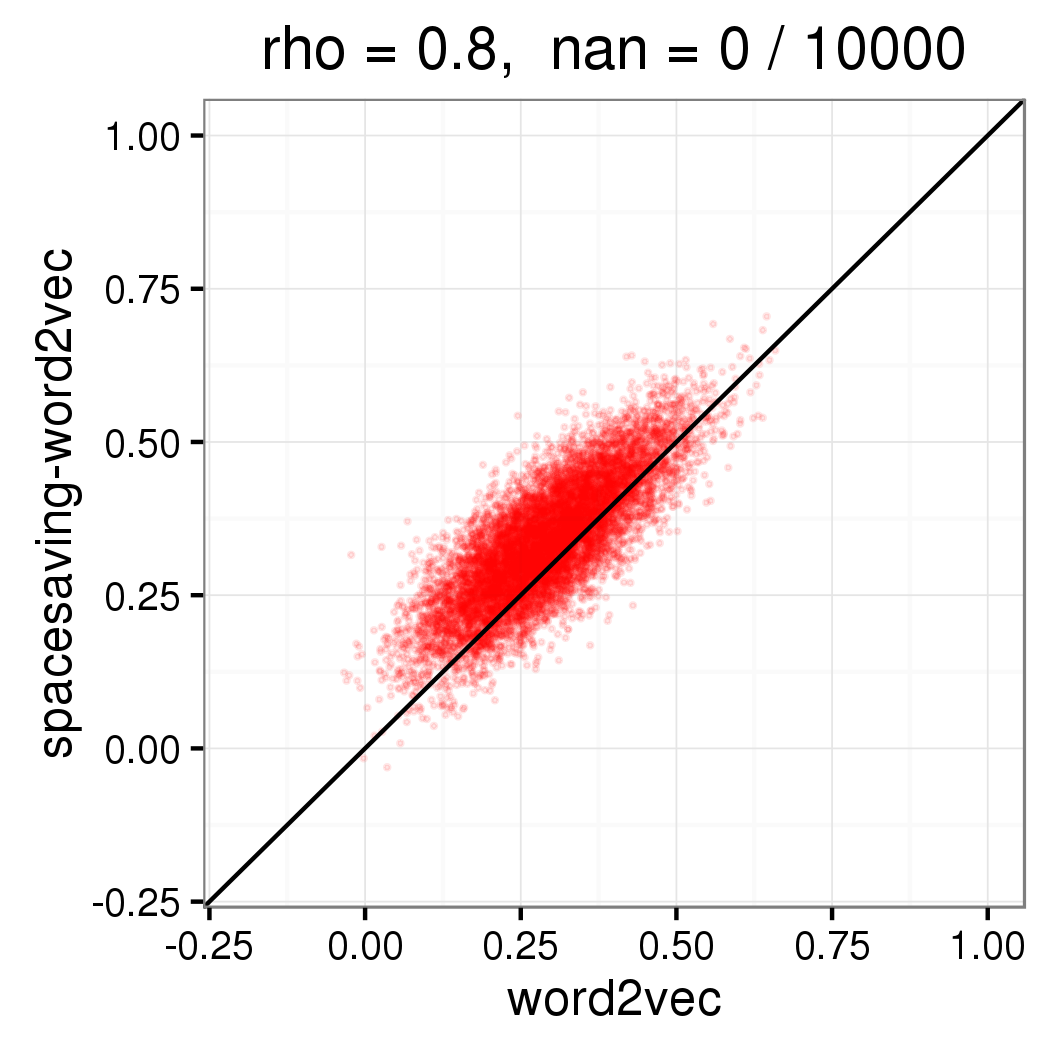}
    \caption{1--100, 1601--1700}
    \end{subfigure}
    \begin{subfigure}[b]{0.49\columnwidth}
    \includegraphics[scale=0.43]{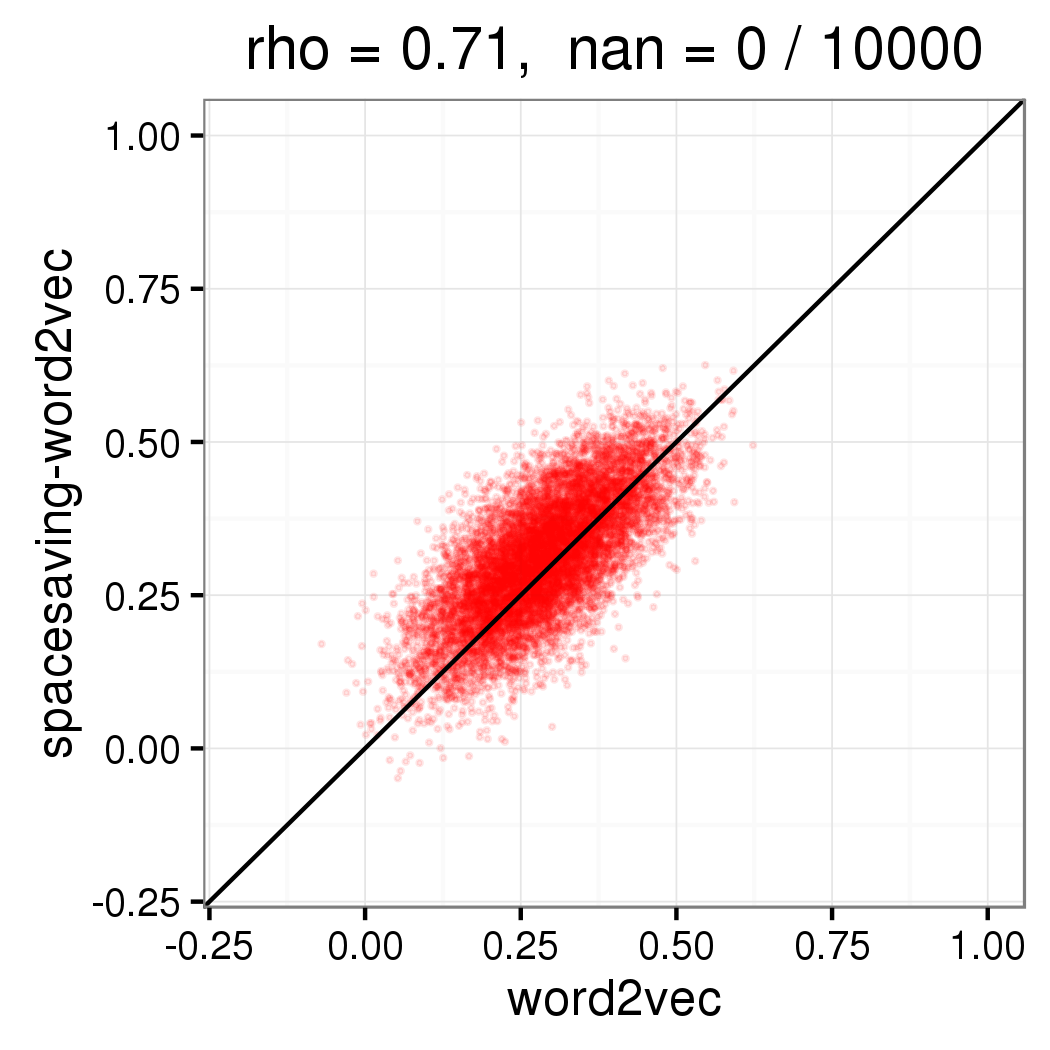}
    \caption{1--100, 6401--6500}
    \end{subfigure}
    \begin{subfigure}[b]{0.49\columnwidth}
    \includegraphics[scale=0.43]{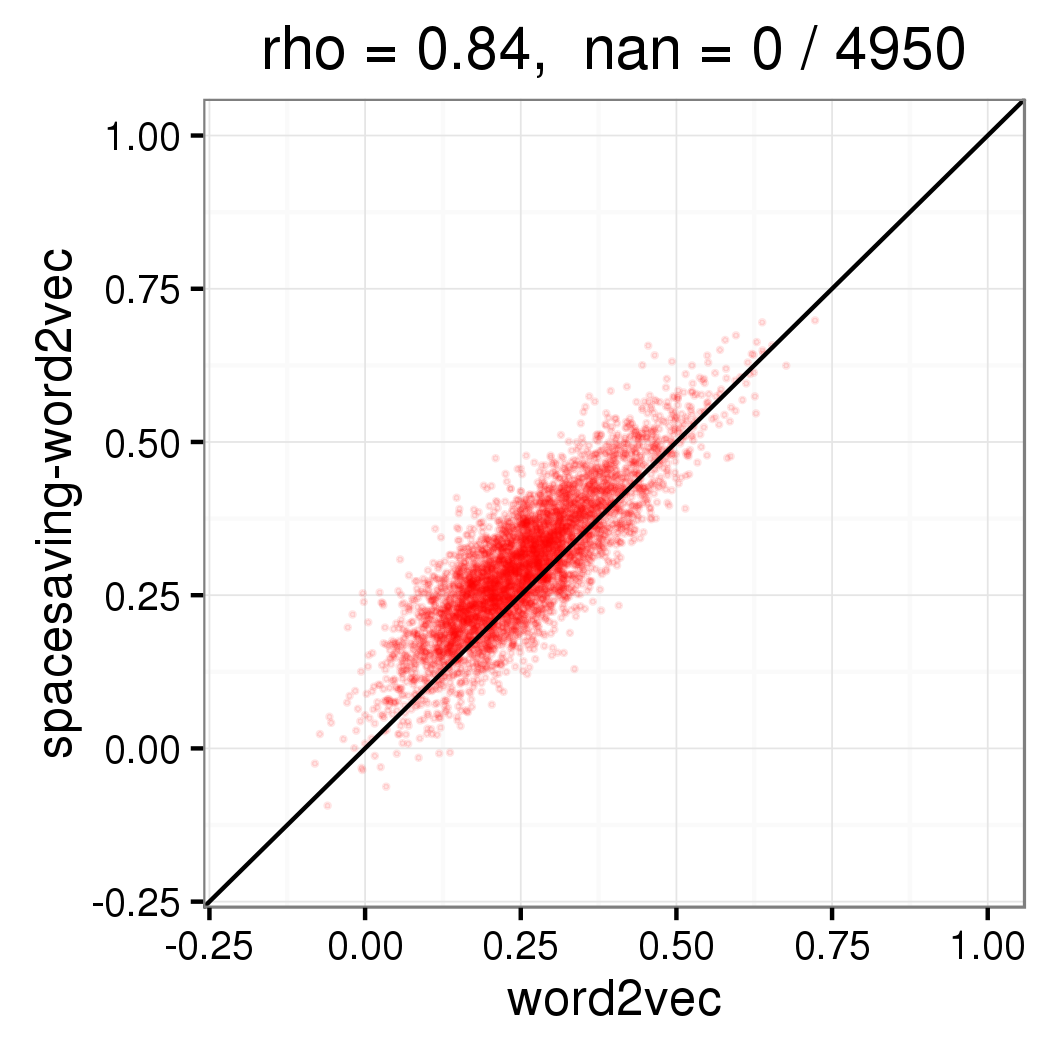}
    \caption{1601--1700, 1601--1700}
    \end{subfigure}
    \begin{subfigure}[b]{0.49\columnwidth}
    \includegraphics[scale=0.43]{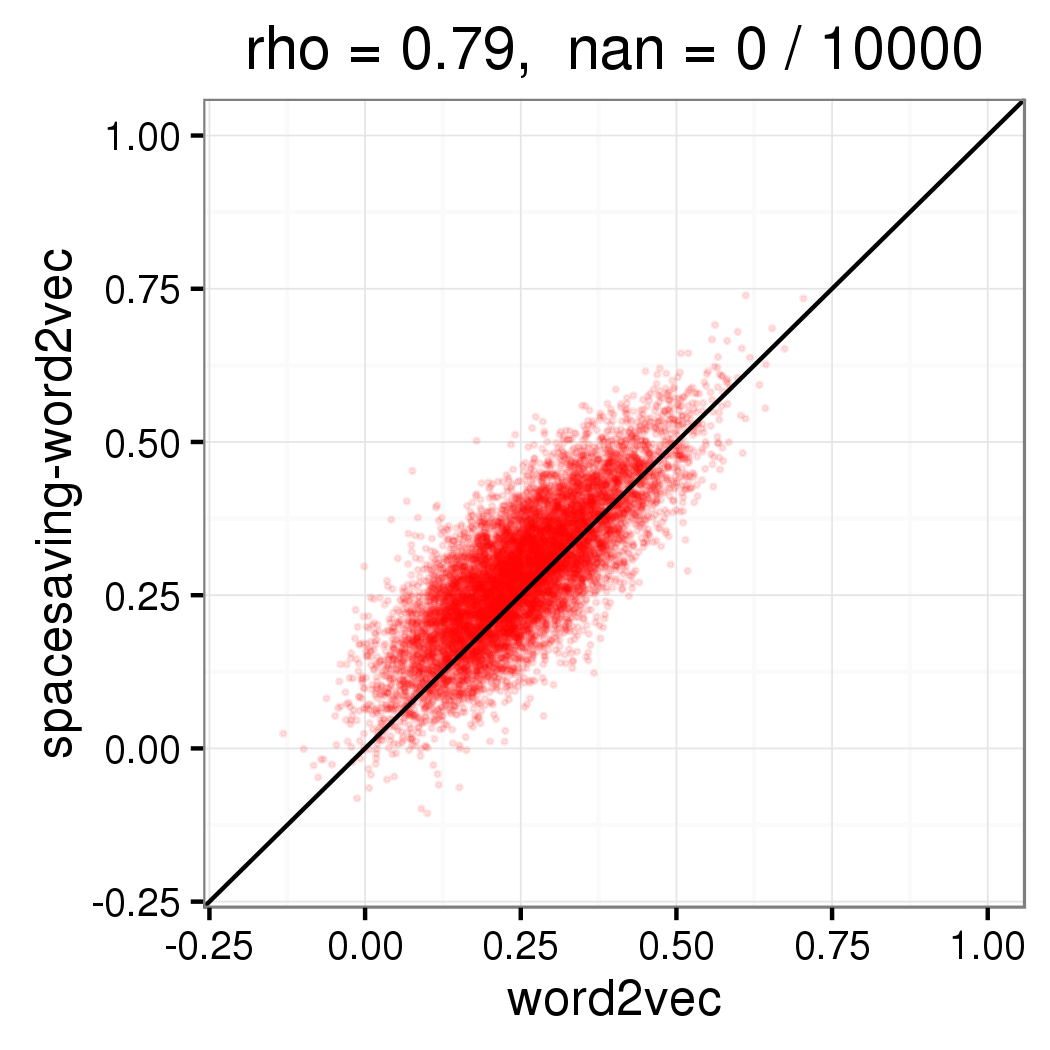}
    \caption{1601--1700, 6401--6500}
    \end{subfigure}
    \begin{subfigure}[b]{0.49\columnwidth}
    \includegraphics[scale=0.43]{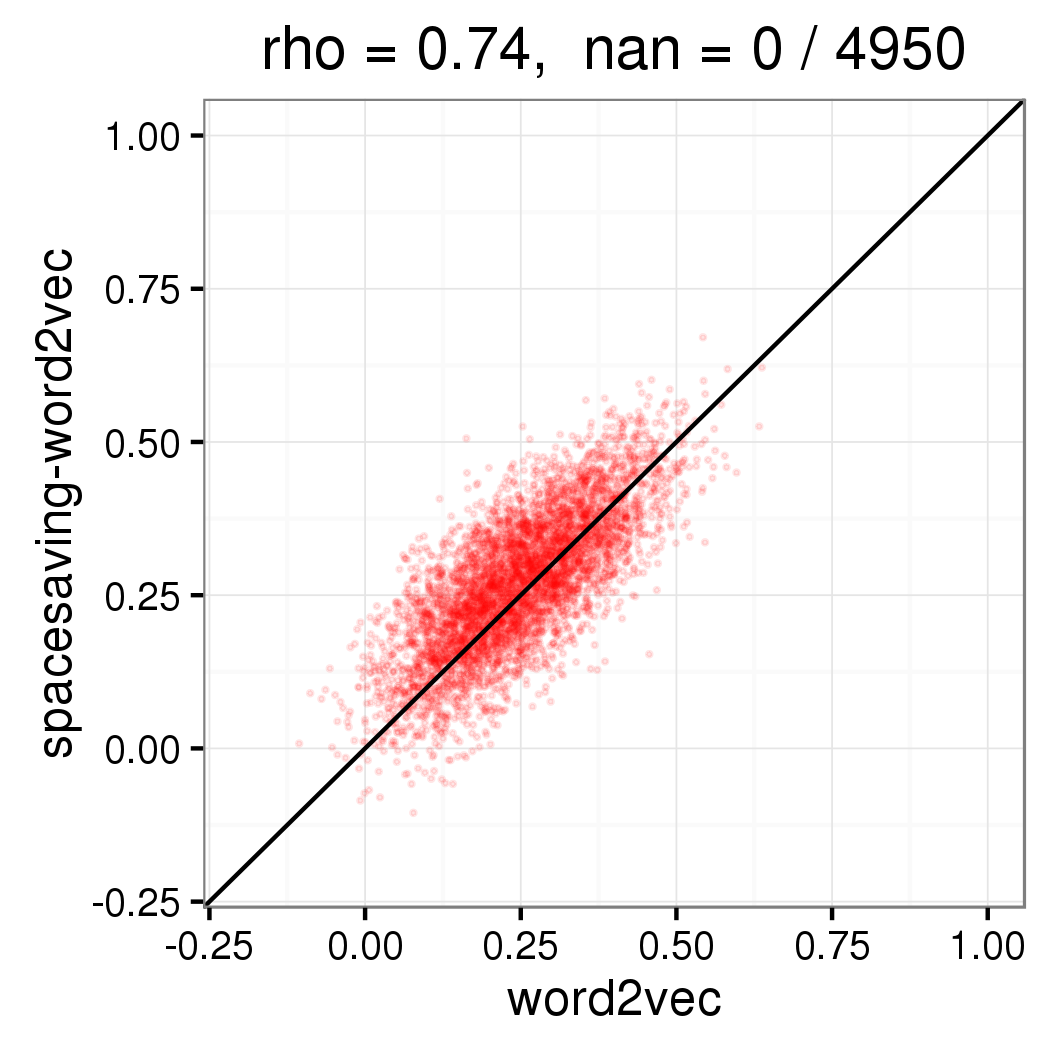}
    \caption{6401--6500, 6401--6500}
    \end{subfigure}
    \caption{Cosine similarity of random word pairs under \sswtv versus
    \wtv, using \num{70000}-dimensional vocabulary.  Models are learned
    on the \texteight data set.}
    \label{fig:text8-70000-sim}
\end{figure}

\begin{figure}
    \scriptsize
    \centering
    \begin{subfigure}[b]{0.49\columnwidth}
    \includegraphics[scale=0.43]{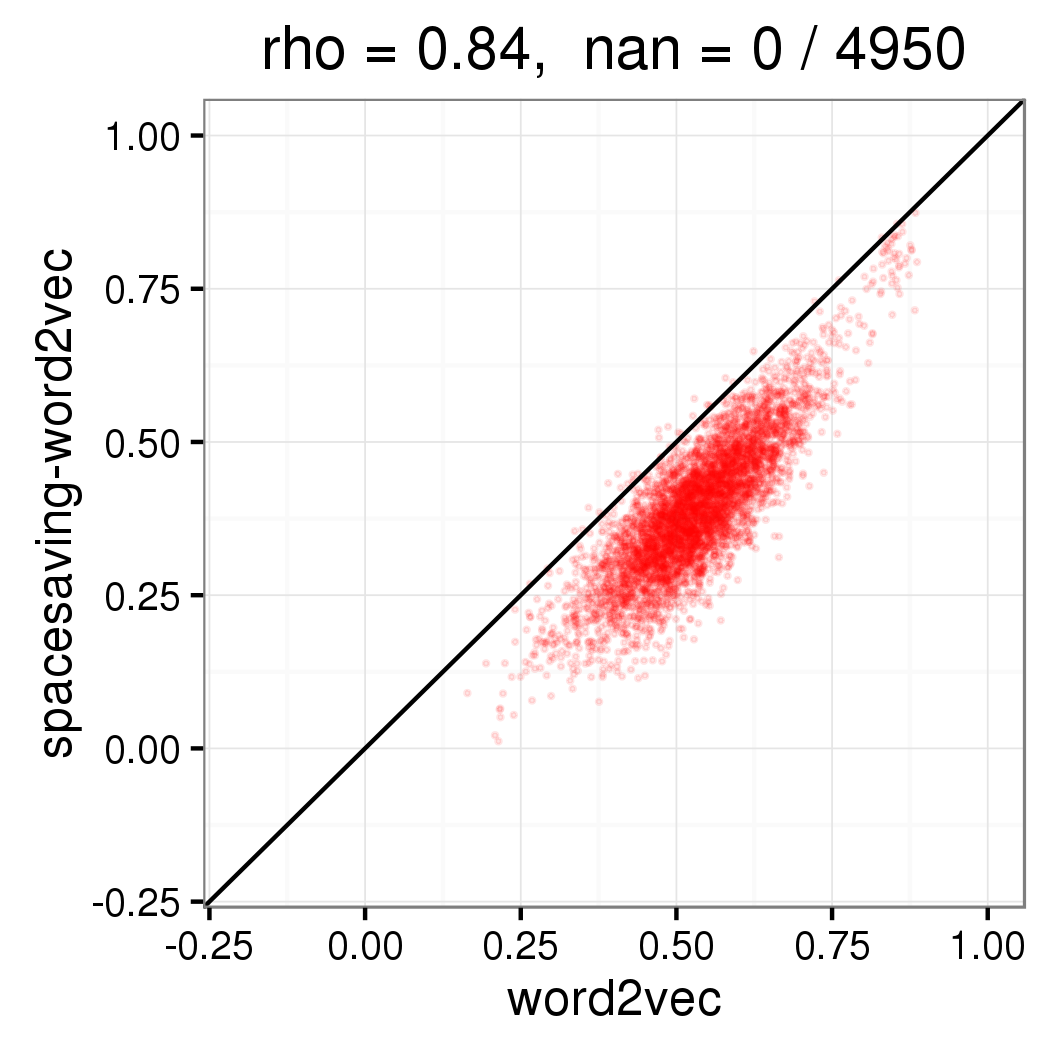}
    \caption{1--100, 1--100}
    \end{subfigure}
    \begin{subfigure}[b]{0.49\columnwidth}
    \includegraphics[scale=0.43]{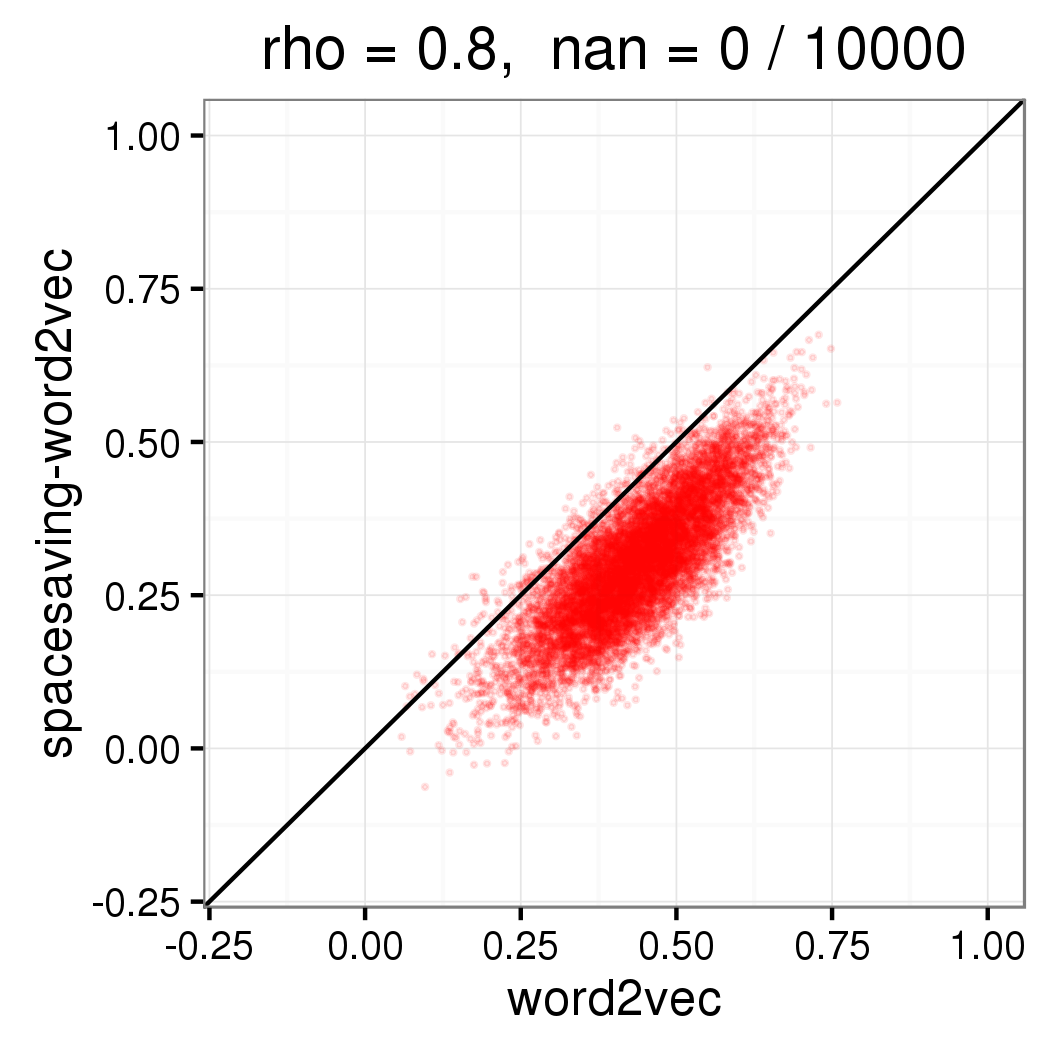}
    \caption{1--100, 1601--1700}
    \end{subfigure}
    \begin{subfigure}[b]{0.49\columnwidth}
    \includegraphics[scale=0.43]{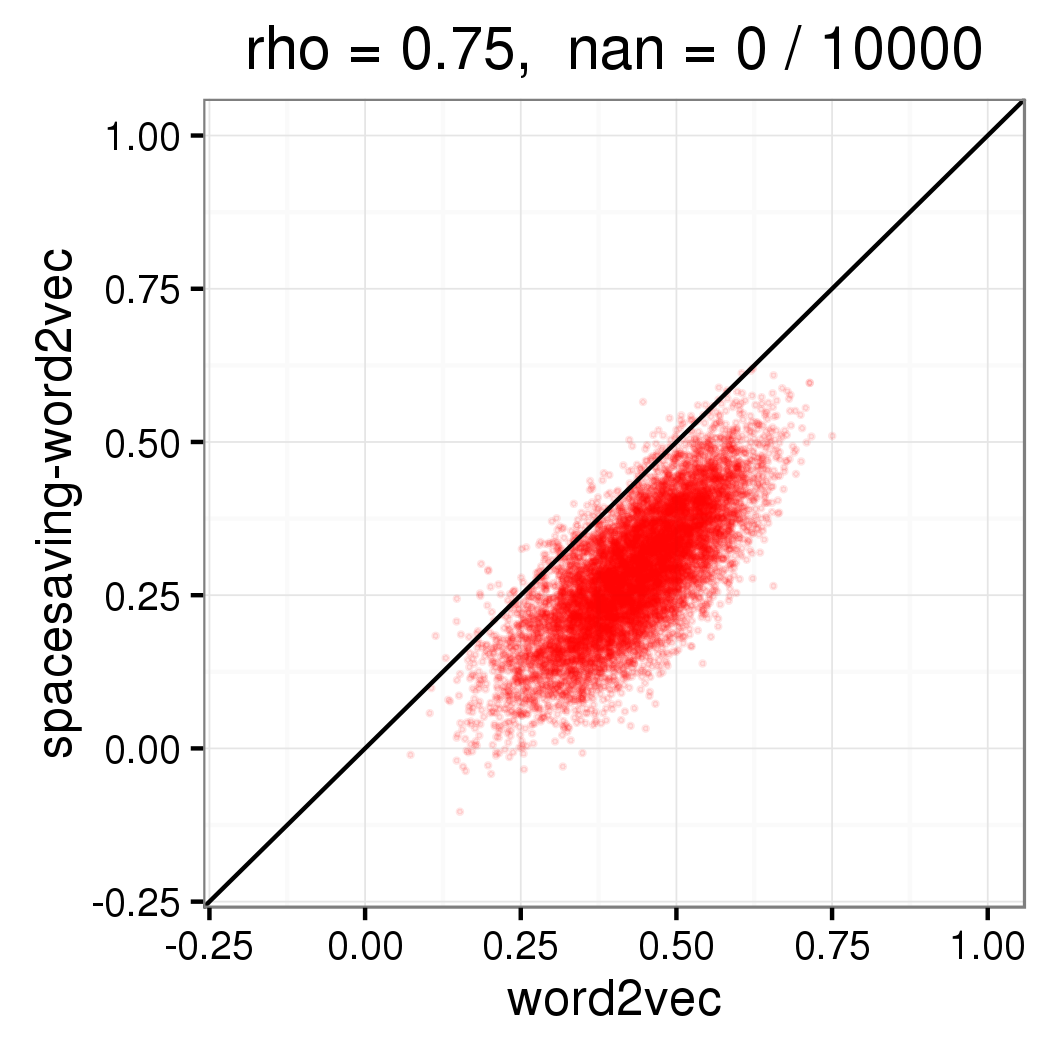}
    \caption{1--100, 6401--6500}
    \end{subfigure}
    \begin{subfigure}[b]{0.49\columnwidth}
    \includegraphics[scale=0.43]{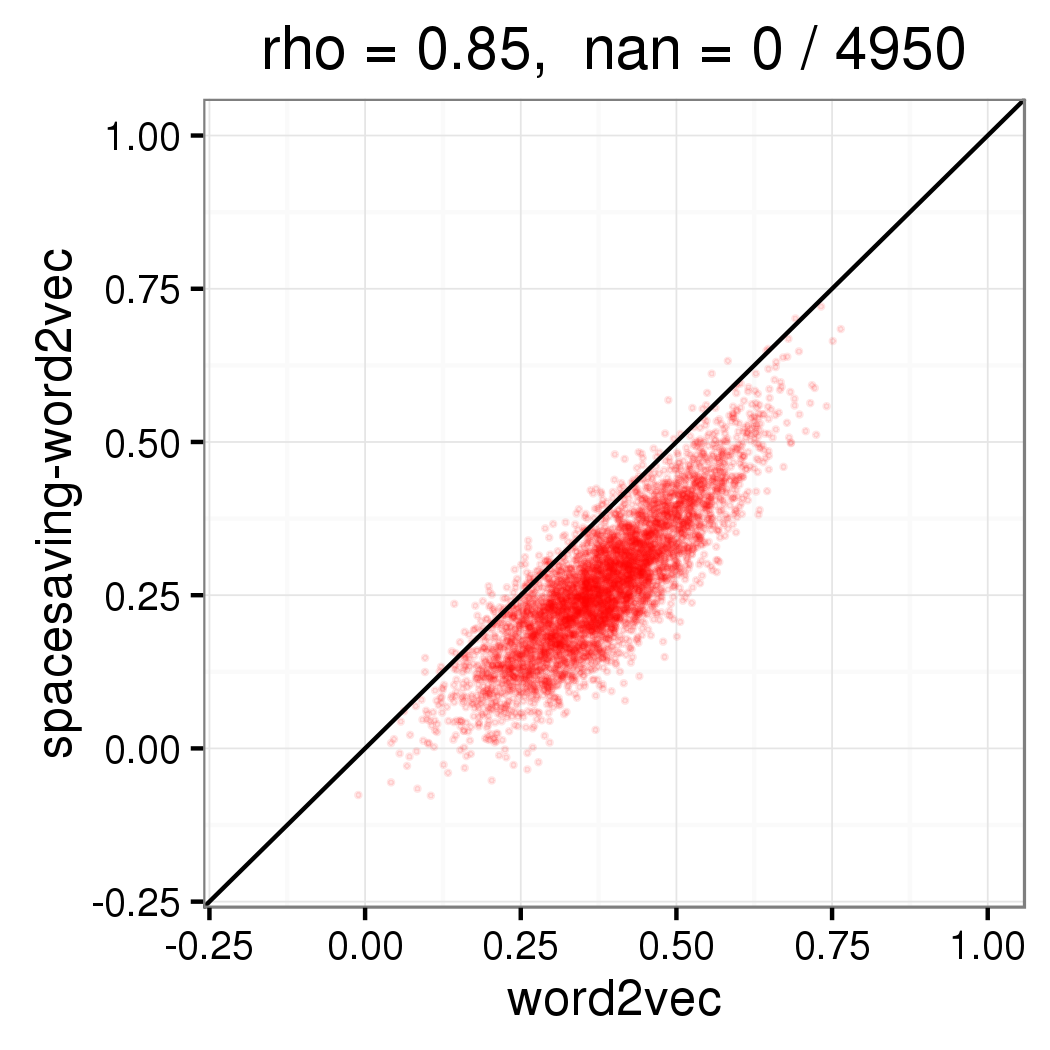}
    \caption{1601--1700, 1601--1700}
    \end{subfigure}
    \begin{subfigure}[b]{0.49\columnwidth}
    \includegraphics[scale=0.43]{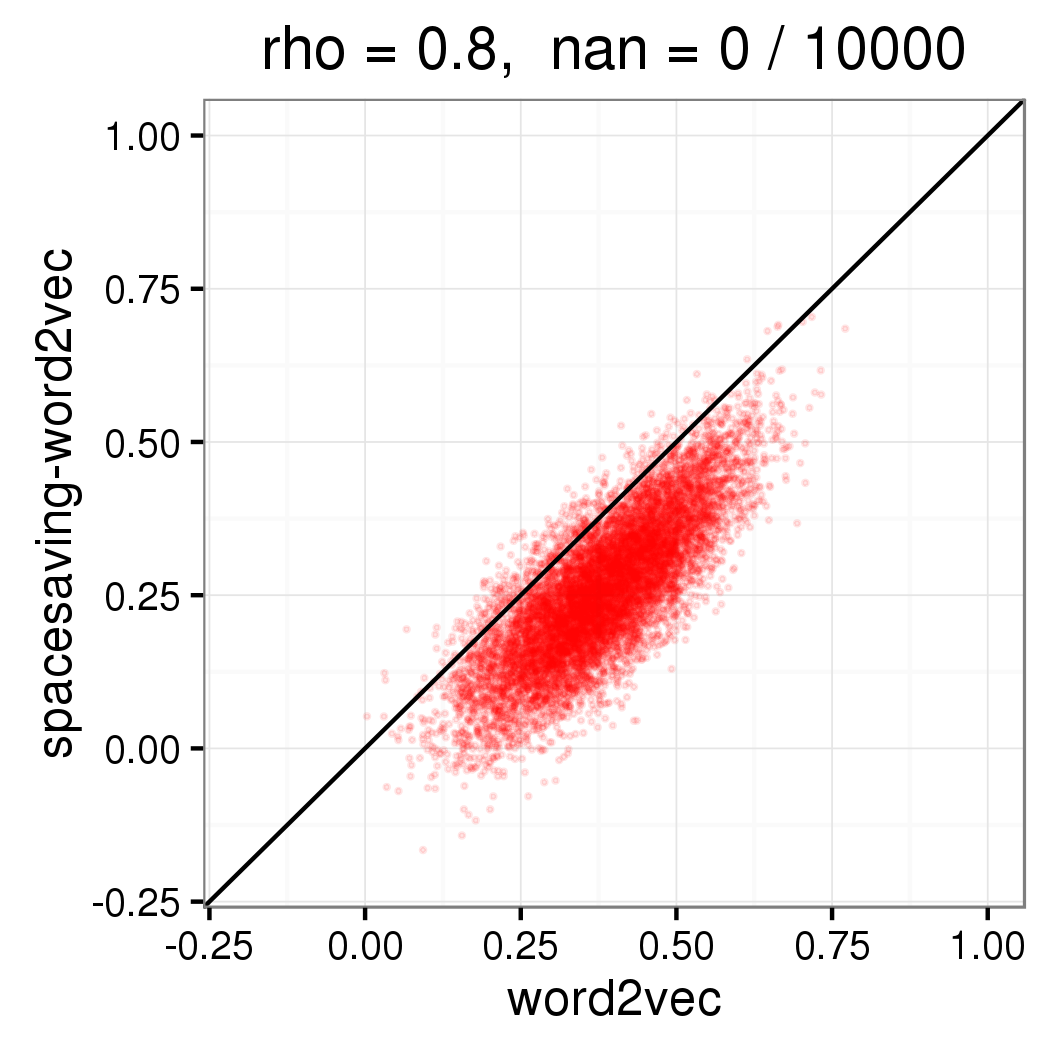}
    \caption{1601--1700, 6401--6500}
    \end{subfigure}
    \begin{subfigure}[b]{0.49\columnwidth}
    \includegraphics[scale=0.43]{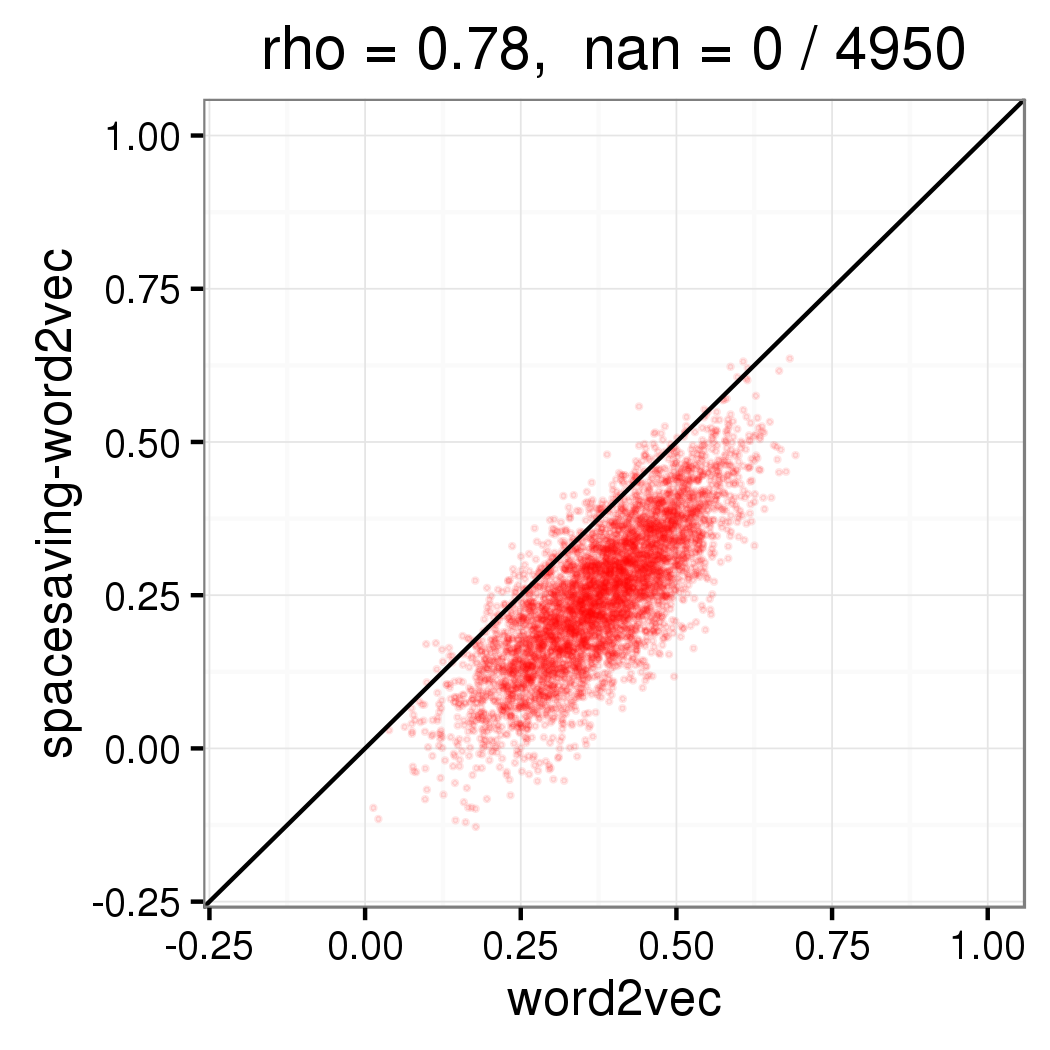}
    \caption{6401--6500, 6401--6500}
    \end{subfigure}
    \caption{Cosine similarity of random word pairs under \sswtv versus
    \wtv, using \num{700000}-dimensional vocabulary.  Models are
    learned on the \texteight data set.}
    \label{fig:text8-700000-sim}
\end{figure}

Next, we assess whether the embeddings learned by \sswtv have similar
pairwise distances as the embeddings learned by \wtv.
We interpret this evaluation as a coarse measurement of how much the
\sswtv embeddings approximate the \wtv embeddings.
As before, we perform this experiment separately for vocabulary sizes
\num{7000}, \num{70000}, and \num{700000}; \wtv and \sswtv are each
trained on \texteight using each vocabulary size in turn.  For each
vocabulary size,
to help illustrate how the \sswtv handles different classes of words,
three 100-word intervals are selected from the true vocabulary ordered
by frequency (descending), namely: words with ranks \num{1} to
\num{100}, words with ranks \num{801} to \num{900}, and words with
ranks \num{6401} to \num{6500}.  For each pair of these intervals, word
pairs (comprising one word from the first interval in the pair and
another word from the second) are drawn uniformly at random and their
cosine similarity under the \sswtv model is plotted against their
cosine similarity under the \wtv model.  These plots are shown in
Figure~\ref{fig:text8-7000-sim} for $K=7000$,
Figure~\ref{fig:text8-70000-sim} for $K=70000$, and
Figure~\ref{fig:text8-700000-sim} for $K=700000$.
Pearson correlation coefficients are reported above each plot; for
$K=70000$ and $K=700000$ we find correlation coefficients in the range
of 0.8 for all pairs of intervals, suggesting that the word
similarities of \sswtv approximate those of \wtv.  For the more
aggressive vocabulary size $K=7000$ the correlation coefficients are
lower, even near zero for low-frequency word intervals.  Moreover, when
the correlation coefficient of the words in the \ssds is near zero the
fraction of word similarities that are undefined due to words being
dropped from the \ssds is also high, near 0.6 or 0.85 depending on the
particular word intervals under consideration.  Considering the true
vocabulary size of \num{253854}, the $K=7000$ case, in which these
deficiencies manifest, is perhaps most
interesting because it yields an appreciable memory savings relative to
the true vocabulary.

Interestingly, the similarities computed by \sswtv are downward-biased
for a trivially large \ssds size of \num{700000}, approximately
unbiased for a non-trivial \ssds size of \num{70000}, and substantially
upward-biased for the aggressive \ssds size of \num{7000}.  We leave
the investigation of the causes and effects of this differential bias
to future work.

\section{Extrinsic Evaluation}
\label{sec:extrinsic}

We now compare the embeddings learned by \wtv and \sswtv in an
experiment designed to test the benefits of the online update afforded
by \sswtv.  In particular,
we apply the learned embeddings in the downstream task of hashtag
prediction~\cite{ding12hashtag,godin13hashtag,weston14tagspace}.
Hashtags are user-defined and user-applied textual labels on
social media posts.
The task of hashtag prediction can be formalized as the
prediction of zero more hashtags from the non-hashtag content of a
post such that the post's author approves of the predicted hashtags for
application to the post.  This task can be operationalized using
historical data by learning to predict the hashtags of a post that were
in fact applied to it.

In our experiment, both \wtv and \sswtv were trained on a sample of
\num{1123701} Tweets from January 2016.  Tweets were tokenized using
the Tift Twitter tokenizer and then normalized by
stripping hashtags, user mentions, and URLs and lower-casing the
remaining text.  In parallel with the \wtv and \sswtv training on Tweet
text, the \ssalg was applied with \num{10000} slots to track the top
(lower-cased) hashtags in the data and a reservoir of size
\num{100000} was used to
maintain a uniform sample of Tweets for each hashtag in that \ssds.
That collection of hashtags was
filtered to the top \num{100} hashtags by frequency, then each
hashtag's reservoir was truncated to \num{1000} samples to reduce skew.
The resulting data set, comprising \num{88413} Tweets, was used to
train the classifiers.  Then, holding the classifiers fixed,
training of the embedding models was resumed
and \num{100}-dimensional hashtag predictions (a binary choice for each
hashtag) were made by the respective classifiers
on a sample of Tweets from February 2016.
For simplicity, predictions were not made on Tweets that did not
contain at least one of the \num{100} hashtags to be predicted; this
simplification resulted in a test set of \num{680579} Tweets.

As we are comparing the \wtv \sgns
implementation to our \sswtv \sgns implementation designed to
mimic it---and considering that we wish to assess performance in the
non-i.i.d.\ streaming setting, in which we cannot practically tune
hyperparameters on the stream and re-train---we used the same
hyperparameters for both methods, choosing the particular values
according to recommendations for \wtv from prior
work~\cite{levy2015improving}.  Specifically, we
used a context window size of two, dynamic context windows,
five negative samples, and subsampling with a threshold of $10^{-3}$.
The most consistently high-performing hyperparameter setting in prior
work was context distribution smoothing of 0.75; we used this value in
\wtv but in \sswtv we sampled words from the unsmoothed empirical
distribution for computational efficiency.

The remaining hyperparameters are the learning rate, the maximum size
of the vocabulary (in \sswtv, the size of the \ssds), the number
of points used to estimate the negative sampling distribution (in
\sswtv, the reservoir size), and the embedding dimension.
Following \wtv, we used a linear learning rate starting at
$2.5 \times 10^{-2}$ and decaying to $2.5 \times 10^{-6}$ over the
first part of the stream, then fixed at $2.5 \times 10^{-6}$ for the
rest of the stream; we used a vocabulary of one hundred thousand;
we used a negative sampling discretization (reservoir)
of one hundred million; and we used an embedding dimension of one
hundred.

The classification problem is formulated as a multi-label
classification task: given a tuple of words representing a Tweet,
transformed to a tuple of vectors by a given embedding model, the task
is to output a 100-dimensional binary prediction in
which 1 represents the presence of a hashtag and 0 represents its
absence.
Note that the classifier
does not update the word vectors (as would be done in supervised
embedding approaches) but merely accepts them from the embedding model
(which is updated in an unsupervised fashion) as input.
Thus we can translate a dynamic and potentially infinite
vocabulary to a static classifier defined on a finite-dimensional
input space.

\begin{figure}
    \scriptsize
    \centering
    \includegraphics[scale=0.43]{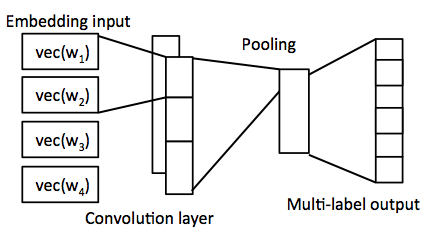}
    \caption{Convolutional neural network classifier for multi-label
    prediction of hashtags given word embeddings as input.  In the
    figure $\mathrm{vec}(w)$ denotes the embedding $v_w$.}
    \label{fig:cnn-architecture}
\end{figure}

Our classifier model uses a convolutional neural network architecture
that was found effective for short text classification in prior
work~\cite{kim14cnn,kurata16multilabel}. Input word
embeddings are convolved along the time dimension, then max-pooled and
passed through a fully connected layer to generate the output vector,
as in Figure~\ref{fig:cnn-architecture}. We use convolution window
lengths of one, two, and three words, and a filtermap size of one
hundred. We train the model using binary
cross-entropy loss~\cite{kurata16multilabel,nam14multilabel}.
We denote the classifier trained on the \wtv model by
\textbf{static-cnn} and the classifier trained on the \sswtv model by
\textbf{stream-cnn}.
The training and testing schemes were
the same for both classifiers; in particular,
both \wtv and \sswtv were updated online during the test phase (on the
February interval of tweets).  However, in that test phase, \wtv marked
all words not in its fixed vocabulary as OOV while \sswtv updated its
vocabulary online.

\begin{table}
\begin{tabular}{|r|r|r|r|}
\hline
embedding/classifier & precision & recall & F1 \\\hline
\textbf{stream-cnn} & 0.40   &   0.49    &  0.42 \\\hline
\textbf{static-cnn} & 0.40   &   0.52    &  0.43   \\\hline
\end{tabular}
\caption{\label{tab:classification_results} Classification results on test
    collection (February 2016).}
\end{table}

At the end of the February
stream we separately measured aggregate classification performance for
\textbf{static-cnn} and \textbf{stream-cnn}.  For each of three
classification metrics, namely
precision, recall, and F1, we computed scores for each of the \num{100}
labels, then aggregated them with a weighted average, weighting by
the number of samples per label in the test set.
The aggregated precision, recall, and F1 scores are shown in
Table~\ref{tab:classification_results}.  We found \textbf{stream-cnn}
achieved an F1 score of 0.42, similar to the F1 score of 0.43 achieved
by \textbf{static-cnn}.  Our original hypothesis was that
\textbf{stream-cnn} would perform better than \textbf{static-cnn} due
to its updated vocabulary; this result does not support that
hypothesis.  This point estimate of the performance difference may
however be interpreted as a sanity check of the \textbf{stream-cnn}
approach, suggesting only a small performance degradation.

\section{Discussion}
\label{sec:discussion}

We first emphasize that both experiments are works in progress and not
meant to provide conclusive validation of \sswtv in their current
forms.
Though the intrinsic experiment measured the error in the cosine
similarities of \sswtv with respect to \wtv, it is not clear what
amount of error (and on what words) is acceptable to maintain a desired
degree of qualitative fidelity in the model; this is a basic matter of
sensitivity analysis.
In particular, it is conceivable that the nearest neighbors of
many words could change even in the presence of
relatively high overall word similarity correlation.
The downstream applications of word embedding models are complex, and
we do not know what sensitivities or invariances they may have without
measuring them.

Meanwhile, though the extrinsic experiment appropriated a real-world
task on real-world data, the particular experimental design is only
illustrative of the motivating application.  Virtually by construction,
the hashtag set defining the target variable of the prediction class
included many hashtags constituting spam, the Tweets of which were
repetitive and arguably simplistic.  The distribution of those hashtags
over the classifier training data was also markedly skewed.
Accordingly, the specific extrinsic experiment reported in this study
might be construed as an artificial spam categorization task, and we
might question whether word embeddings and a convolutional neural
network classifier are necessary to perform the task in the first
place.  The extrinsic experiment requires substantial
refinement in order to reflect the natural real-world hashtag
prediction task, and a closer analysis of results (including
comparison against strong baselines) is necessary before the
validation potential of the experiment can be known.

There were several design choices in the extrinsic
experiment on Twitter.
These included the choice of time span of Tweets to use for
initialization of the embedding models and for prediction by the
classifier; the frequency at which predictions were made by the
classifier (for example, predicting after every Tweet, after every $B$
Tweets, or after all Tweets in the prediction set); the filtering,
normalization, and segmentation of the Tweets;
the construction of the set of hashtags to be predicted;
the treatment of Tweets with no hashtags;
the choice to update \sswtv and \wtv at test time rather than keeping
one or both embedding models fixed; and
the estimation by each embedding algorithm of its own vocabulary
(rather than an ablation analysis in which \wtv was seeded with the
overall \sswtv vocabulary).
Our solutions to these choices are by no means the best solutions in
terms of most thoroughly evaluating \sswtv against \wtv,
but for the sake of scope we must defer more in-depth investigation to
future work.

In particular,
we did not analyze the dynamics of \sswtv on non-stationary data in our
experiments: though the extrinsic experiment used a real-world
non-stationary data stream, we truncated the stream to a relatively
short interval and we only measured aggregate performance at the end of
that interval.
Therefore we must leave empirical testing of the
ability of \sswtv to accommodate a highly dynamic stream, including a
stream exhibiting sudden shifts in distribution, to future work.  In
particular, if embeddings of common words that stay in the \ssds rotate
during training, those word embeddings would not be useful in
off-the-shelf downstream models, which often expect the embedding of
a given word to be static.

Future work may benefit from studying the relationship between context
distribution
smoothing, in which the negative sampling distribution is made a
smoothed version of the empirical unigram language model, and
subsampling, in which words observed by the embedding model are
subsampled from the input according to frequency.  These investigations
are particularly interesting in the streaming setting, in which context
distribution smoothing is computationally expensive whereas subsampling
is easy.

Though prior work found context distribution smoothing consistently
beneficial~\cite{levy2015improving}, in the preliminary experiments
reported in this study we have not seen such a strong preference.
Though this lack of confirmation may be due to the rudimentary nature
of our experiments, we speculate that the over-counting of
low-frequency words by the \ssalg,
and resultant over-representation of low-frequency words in the
negative sampling distribution, may constitute context distribution
smoothing as a happy side effect.

The frequency of ejections in the tail of the \ssds impact not
only runtime but the utility of the learned embeddings, as
rapid ejections could result in a word towards the beginning of a
sentence being ejected before the last word of the sentence is added
to the \ssds and the embedding model is updated.  However, if the
\ssds has a long tail, this event may occur only rarely.
That is, if there is a long tail, after a
target item is inserted in the \ssds a large number of subsequent
items (the items with lowest count) must be inserted before the target
item has a chance of being ejected.  It may be useful to study the
empirical frequency and impact of ejections of recently inserted words
in future work.

The \sswtv algorithm handles an unrecognized word at training time by
adding it to the \ssds, ejecting an existing word and resetting its
embedding, and updating the new embedding based on the context of the
unrecognized word.  (Subsequent contexts of that word are then used to
further update that embedding, assuming the word is not ejected from
the \ssds in the meantime.)  This approach treats each word type
atomically and is thus agnostic to morphological structure; it cannot
infer embeddings for
unrecognized words out of context, and may yield a low-quality
embedding for an unrecognized word the first time (or first few times)
it is seen.  We leave the measurement and resolution of these issues to
future work; it may be beneficial in particular to back off to sub-word
unit representations.

The algorithm proposed here is complementary to incremental
SGNS, sharing a similar motivation and high-level
approach while exhibiting several different implementation choices.
We developed our algorithm independently, and do not compare it
empirically to incremental SGNS, but doing so would be an interesting
direction for future work.  We moreover suggest an ablation study,
evaluating each design decision in isolation in order to better
understand the empirical operation of these algorithms and perhaps
develop a superior third implementation by combining the
best-performing components from \sswtv and incremental SGNS.

In incremental SGNS, the modified reservoir sampling approach to
estimating the smoothed negative sampling distribution either requires
memory linear in the true vocabulary (in order to maintain cumulative
weights of all words seen so far) or suffers an approximation error in
computing the cumulative weights of low-probability words.  Our
approach faces the same memory-bias trade-off as incremental SGNS, and
we choose to achieve constant memory usage while over-estimating the
frequency of low-probability words in the negative sampling
distribution.  A second source of bias is introduced in incremental
SGNS by deterministically making the expected number of insertions into
the reservoir, instead of sampling the number of insertions, for
computational efficiency.  The \sswtv negative
sampling distribution is not smoothed, hence we afford constant-time
reservoir updates without incurring this additional source of
bias.\footnote{
    Incidentally, the original \wtv implementation uses a deterministic
    construction of the negative sampling table and is also biased with
    respect to the smoothed empirical unigram distribution.
}

Compellingly,
the optimal solution of incremental SGNS is shown to have an objective
value under the batch SGNS objective function that converges in
probability to the optimal batch SGNS objective value, given
i.i.d.\ data~\cite{kaji2017}.  (However, the incremental SGNS objective
function in this analysis employs an unbiased incremental negative
sampling distribution, and so differs from the objective function that
is implemented.)
Empirically, when the effective vocabularies are constrained to be
similar in size, incremental SGNS emulates the performance of batch
SGNS and \wtv on semantic similarity and word analogy tasks.
Updating a pre-trained incremental SGNS model on new data also yields a
significant runtime improvement over re-training a batch SGNS model (or
\wtv) on the combined old and new data.

We would be remiss not to comment on the practical performance of our
approach.  Despite implementing \sswtv in C++ and closely
following the tricks employed by \wtv, in informal experiments we found
\wtv consistently faster by a factor of two or more in the main loop
(after computing the vocabulary and negative sampling distribution),
while the time taken by \wtv to compute the vocabulary and negative
sampling distribution was relatively small.  Indeed, the practical
performance of \wtv is quantified by the runtime experiments comparing
incremental SGNS to its batch version and \wtv in prior
work~\cite{kaji2017}.  Specifically, the three algorithms are compared
in runtime on
the task of updating a pre-trained model based on new data.  For batch
SGNS and \wtv this is operationalized as re-training the model on both
old and new data sets, while incremental SGNS is run only on the new
data set.  When the old data set comprises ten million words and the
new data set comprises one million, incremental SGNS is found to
achieve an overall speed-up of 7.3 over \wtv.  However, in this
experiment incremental
SGNS is processing one-eleventh as much data; it is thus only
two-thirds as
fast as \wtv \emph{per word}.  Moreover, while batch SGNS and \wtv each
scan the combined (old and new) data set twice in that
experiment, effectively corresponding to 22 scans of the new
data set, incremental SGNS only scans the new data set once.  We do
not argue this is a shortcoming in the implementation of incremental
SGNS; on the contrary, this performance gap reflects our experience
with our own implementation and appears to point to clever
manual optimization evident in the \wtv code.  We therefore caution
practitioners and implore other researchers to carefully consider
whether the theoretical benefits of streaming algorithms such as these
are nullified by practical inefficiencies.

On the note of applications,
our extrinsic experiment used hashtag prediction as a motivating
downstream task.
A hashtag prediction model can be used to suggest hashtags
to users writing new Tweets. It could also be used to provide
additional (inferred) data to downstream analytics. However, this
latter application raises a dual-use concern: if a user
intentionally refrains from using a hashtag in order to escape
categorization or publicity, a hashtag prediction model could
reduce or remove that particular form of privacy potentially without
the user's consent.  The technology developed in the current work,
enabling embeddings to be learned at greater scale, could thus be used
to help a user or to harm them.

\section{Conclusion}
\label{sec:conclusion}

We have developed a one-pass, bounded-memory variant of the popular
\sgns algorithm for training word embeddings.
Our approach, called \sswtv after the \wtv implementation of
\sgns on which it is based, leverages the \ssalg and reservoir
sampling in order to maintain an approximate, dynamic vocabulary and
negative sampling distribution.
Though preliminary experiments provide some evidence for the fidelity
of \sswtv to \wtv, there are still many open questions to be addressed.
While we cannot yet wholeheartedly endorse the use of \sswtv in
real-world applications, we hope that thoughtful future research
drawing on insights gleaned from \sswtv and incremental
SGNS~\cite{kaji2017} will close this gap.


\bibliography{references}
\bibliographystyle{acl2012}

\clearpage

\appendix
\section{Appendix}

The complete \sswtv learning algorithm is listed in
Algorithm~\ref{alg:sswtv-full}.

\begin{algorithm*}
    \KwData{Stream of sentences $(s_i)_i$,
    subsampling threshold $\delta$,
    vocabulary size $K$, negative sampling reservoir size $N$,
    negative sample size $S$,
    embedding dimension $D$, context radius $C$,
    learning rate offset $\tau$, learning rate exponent $\kappa$.}

    \KwResult{Any-time word embeddings $(v_k \colon k \in [K])$
    indexed against \ssds $\phi \colon [K] \to
    \Sigma^*$.}

    initialize empty size-$K$ \ssds and size-$N$ negative sampling
    reservoir \;

    initialize input, output word embeddings
    ($v_k \sim \Normal(0, 1)^D$, $v'_k \sim \Normal(0, 1)^D$)
    for all $k \in [K]$ \;

    initialize $t_k \gets 1$ for all $k \in [K]$ \;

    \For{ever}{
        read sentence $s = (w_1, \ldots, w_J)$ (a tuple of words) \;
        subsample sentence $s' = (w'_1, \ldots, w'_{J'})$ \;
        \For{$1 \le j \le J'$}{
            insert $w'_j$ into \ssds and its index into
            reservoir \;
        }
        \For{$1 \le j \le J' - 2 C$}{
            $m \gets j + 2 C$ \;
            \tcp{$j$ is start of current context; $m$ is end}

            \If{all words $w'_j, \ldots, w'_m$ are in \ssds}{
                let $k_j, \ldots, k_m$ be \ssds indices of $w'_j,
                \ldots, w'_m$ \;

                \tcp{iterate over output word $w'_\ell$}
                \For{$j \le \ell \le m$}{

                    \tcp{$w'_{j+C}$ is the input word}
                    \If{$\ell \ne j + C$}{

                        \tcp{take gradient step on output word}

                        $\alpha \gets 1 - \sigma(\langle v_{k_{j+C}},
                        v'_{k_\ell}\rangle )$ \;

                        $v'_{k_\ell} \gets v'_{k_\ell} +
                        \rho_{k_\ell} \alpha v_{k_{j+C}}$ \;

                        $u \gets \rho_{k_{j+C}} \alpha v'_{k_\ell}$ \;

                        \tcp{take gradient steps on negative samples}
                        \For{$1 \le j' \le S$}{
                            draw $k_{(-)}$ from negative sampling
                            reservoir \;

                            $\alpha \gets - \sigma(\langle v_{k_{j+C}},
                            v'_{k_{(-)}}\rangle )$ \;

                            $v'_{k_{(-)}} \gets v'_{k_{(-)}} +
                            \rho_{k_{(-)}} \alpha v_{k_{j+C}}$ \;

                            $u \gets u + \rho_{k_{j+C}} \alpha v'_{k_{(-)}}$ \;
                        }

                        \tcp{take gradient step on input word}
                        $v_{k_{j+C}} \gets v_{k_{j+C}} + u$ \;

                        \For{indices $k$ of union of input, output,
                        negative sample words in this context}{
                            $t_k \gets t_k + 1$ \;
                        }
                    }
                }
            }
        }
    }
\caption{Complete \sswtv algorithm.}
\label{alg:sswtv-full}
\end{algorithm*}

\end{document}